\documentclass[10pt,twocolumn,letterpaper]{article}

\usepackage[pagenumbers]{cvpr} 

\definecolor{cvprblue}{rgb}{0.21,0.49,0.74}
\usepackage[pagebackref,breaklinks,colorlinks,allcolors=cvprblue]{hyperref}

\usepackage{xcolor}
\usepackage{microtype}
\usepackage{amsmath}
\usepackage{amsfonts}
\usepackage{subcaption}
\usepackage{float}
\usepackage{dblfloatfix}
\usepackage{multirow}

\title{VOST-SGG: VLM-Aided One-Stage Spatio-Temporal Scene Graph Generation}

\author{Chinthani~Sugandhika$^{1,2,3}$~\and~Chen~Li$^{2,3}$~\and~Deepu~Rajan$^{1}$~\and~Basura~Fernando$^{1,2,3}$ \\
$^{1}$College of Computing and Data Science, Nanyang Technological University, Singapore\\
$^{2}$Institute of High-Performance Computing, Agency for Science, Technology and Research, Singapore\\
$^{3}$Centre for Frontier AI Research, Agency for Science, Technology and Research, Singapore
}

\begin{document}
\maketitle

\begin{abstract}


Spatio-temporal scene graph generation (ST-SGG) aims to model objects and their evolving relationships across video frames, enabling interpretable representations for downstream reasoning tasks such as video captioning and visual question answering. Despite recent advancements in DETR-style single-stage ST-SGG models, they still suffer from several key limitations. First, while these models rely on attention-based learnable queries as a core component, these learnable queries are semantically uninformed and instance-agnostically initialized. Second, these models rely exclusively on unimodal visual features for predicate classification. To address these challenges, we propose VOST-SGG, a \textbf{V}LM-aided \textbf{O}ne-stage \textbf{ST-SGG} framework, that integrates the common sense reasoning capabilities of vision-language models (VLMs) into the ST-SGG pipeline. First, we introduce the dual-source query initialization strategy that disentangles what to attend to, from where to attend, enabling semantically grounded “what-where” reasoning. Furthermore, we propose a multi-modal feature bank that fuses visual, textual, and spatial cues derived from VLMs for improved predicate classification. Extensive experiments on the Action Genome dataset demonstrate that our approach achieves state-of-the-art performance, validating the effectiveness of integrating VLM-aided semantic priors and multi-modal features for ST-SGG. We will release the code at \href{https://github.com/LUNAProject22/VOST}{https://github.com/LUNAProject22/VOST}.

\end{abstract}    
\section{Introduction}
\label{sec:intro}

\begin{figure}[ht]
    \centering
    \begin{subfigure}[]{1.0\linewidth}
        \centering
        \includegraphics[width=\linewidth]{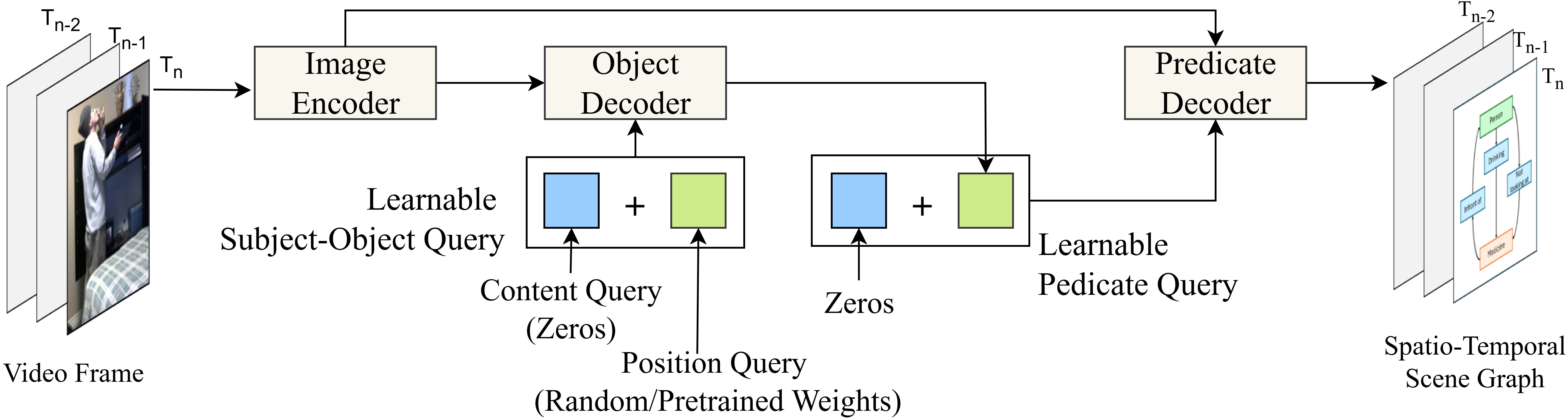}
        \subcaption{Previous work.}
        \label{fig:intro_diagram_oed}
    \end{subfigure}
    \hfill
    \begin{subfigure}[]{1.0\linewidth}
        \centering
        \includegraphics[width=\linewidth]{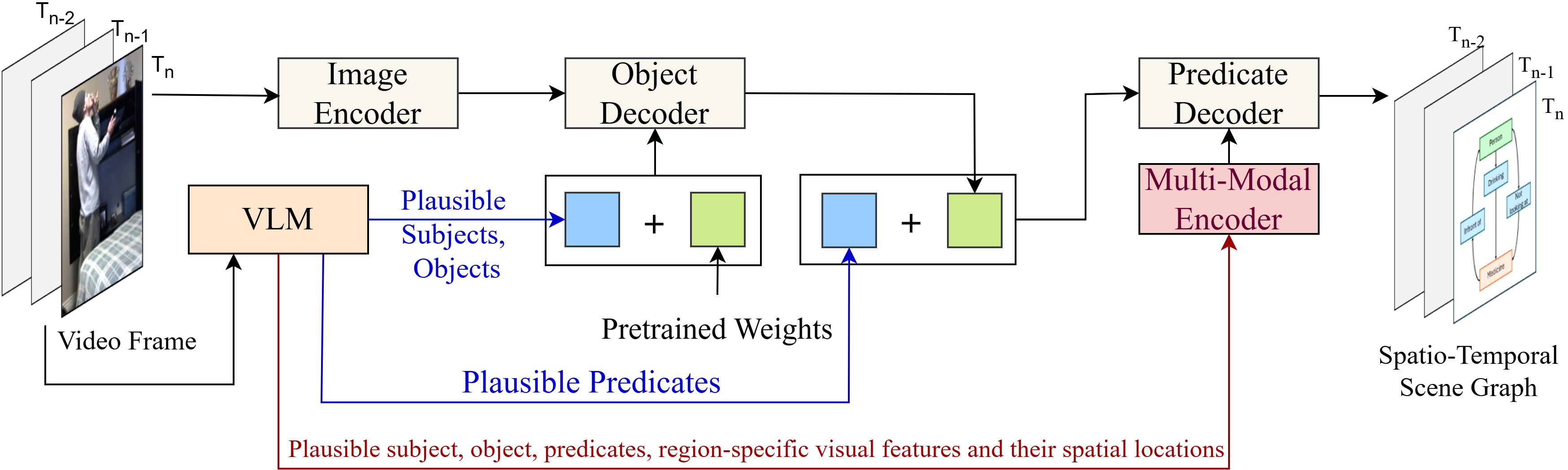}
        \subcaption{VOST-SGG (Ours).}
        \label{fig:intro_diagram_our}
    \end{subfigure}
    \caption{Comparison between (a) Previous work and, (b) Our proposed VOST-SGG framework.}
    \label{fig:intro_diagram}
\end{figure}


Scene Graph Generation (SGG) is a core computer vision task that transforms visual content into structured graphs, where nodes represent objects and edges capture their relationships. This structured representation has become essential for high-level visual–semantic reasoning tasks such as image captioning \cite{gao2018image}, visual question answering \cite{hudson2019gqa, wu2024star}, and image retrieval \cite{wang2020cross}. While SGG for static images has has seen considerable progress, extending these capabilities to video, termed Spatio-Temporal Scene Graph Generation (ST-SGG), is gaining popularity in recent years. With the rapid growth of video data in fields like surveillance and robotics, improving ST-SGG is crucial for achieving fine-grained, semantically meaningful machine understanding of dynamic visual scenes.

Early approaches to ST-SGG adopt a multi-stage pipeline comprising instance detection, temporal association, and relation classification \cite{sttran, apt}. 
However, these pipelines rely on independently trained modules, which disrupts joint optimization and limits cross-task synergy. 
Recent research has explored one-stage ST-SGG models \cite{oed, tpt}, which unify these steps within an end-to-end Transformer-based architecture inspired by DETR \cite{detr}. 
In DETR-like models, learnable queries play pivotal role in guiding the attention mechanism of the Transformer decoder. 
Each learnable query comprises two components: a \emph{content} query, which specifies what the model should focus on, and a \emph{position} query, which indicates where to attend \cite{yao2021efficient}. The cross-attention weights are computed by comparing a query with a set of keys which also consists  of two parts as a content part (encoded image feature) and a positional part (positional embedding). As highlighted by \citet{liu2022dab}, this design enables the decoder to effectively pool features from a feature map based on the query-to-feature similarity.

Existing one-stage DETR-like ST-SGG models still exhibit several limitations: 
\textbf{(1) Lack of semantic specificity in query formulation}: Although content queries are intended to encode what to look for, prior work typically assign them a zero vector (see Fig.~\ref{fig:intro_diagram_oed}), offering no semantic cues regarding the objects or predicates to retrieve. Instead, these methods focus mainly on the spatial component, i.e. "where" to look, by initializing position queries with random or pre-trained weights;
\textbf{(2) Instance-agnostic query formulation}: Queries are typically formulated without considering the specific visual/semantic context of the video frame, limiting their ability to focus on frame-specific relationships;
\textbf{(3) Uni-modal predicate representation}: These models rely solely on visual features, which hinders their ability to disambiguate semantically rich relationships  
that may appear visually similar but differ in functional meaning. 
Disambiguating such interactions often requires contextual or common sense knowledge that goes beyond visual appearance.

To address these limitations, we introduce \textbf{VOST-SGG}, a novel \textbf{V}LM-aided \textbf{O}ne-stage \textbf{ST-SGG} framework that integrates the common sense reasoning capabilities of Vision-Language Models (VLMs) into one-stage ST-SGG pipeline via \textbf{two key architectural innovations:}
\textbf{(1) Dual-source query initialization:} To overcome the first two limitations, we infuse instance-specific high-level common sense knowledge obtained from VLMs into the content queries allowing the decoder to \emph{look for} semantically meaningful objects and predicates rather than a zero vector as shown in Fig. \ref{fig:intro_diagram_our}. Further, position queries are initialized using instance-agnostic spatial anchors pre-trained on MS-COCO \cite{tpt}, providing spatial stability and adaptability across varying scenes. 
\textbf{(2) Multi-modal feature bank for predicate decoding}: To mitigate the third limitation, we introduce a multi-modal feature bank that uses visual, textual and spatial features. Specifically, these features are generated from a VLM generated textual cues for subjects, objects, and predicates with their region-level visual embeddings and spatial information. During decoding, learnable queries cross-attend over this multi-modal features space instead of relying solely on image features, enabling more robust and context-aware predicate classification.

In summary, we present a VLM-aided one-stage ST-SGG framework overcoming three  limitations of existing one-stage ST-SGG models. Our contributions are three-fold:
    (1) We revisit the formulation of learnable queries in DETR-style ST-SGG decoders and propose a \textbf{dual-source query initialization strategy} that disentangles what to attend to (with instance-specific and semantically meaningful content queries) from where to attend (with instance-agnostic position queries), facilitating more expressive \textbf{“what-where” reasoning} in decoder.
    (2) We introduce the concept of \textbf{multi-modal feature bank}, that fuses visual, textual, and spatial cues derived from the common sense knowledge of VLMs, offering richer contextual information for improved predicate classification.
    (3) Through extensive experiments on Action Genome dataset \cite{ag}, we demonstrate that our approach achieves state-of-the-art performance, validating the effectiveness of our VOST-SGG framework for spatio-temporal scene graph generation.
\section{Related Work}
\label{sec:previous_work}

\paragraph{Multi-stage spatio-temporal scene graph generation. }
Similar to how image-based scene graph generation gained significant attention with the introduction of the Visual Genome dataset \cite{krishna2017visual}, ST-SGG has attracted interest following the release of the Action Genome dataset \cite{ag}. Building on the two-stage paradigm of image-based SGG that comprises of stages of object detection using off-the-shelf detectors followed by predicate classification \cite{xu2017scene, yang2018graph}, multi-stage ST-SGG models introduced an additional step for temporal modelling.
However, these methods rely on multiple specialized modules \cite{tempura, tr2}, requiring separate training schemes that hinder joint optimization and cross-task synergy. Moreover, the exhaustive enumeration of subject-object pairs results in significant computational overhead and redundancy.

\paragraph{One-stage spatio-temporal scene graph generation. }
One-stage ST-SGG models \cite{oed, zheng2022vrdformer}, inspired by the success of DETR-style \cite{detr} image-based scene graph generation \cite{cong2023reltr}, have recently emerged to overcome the drawbacks of multi-stage pipelines. These approaches unify object detection, temporal context modeling, and predicate classification within a single end-to-end framework, enabling joint optimization. 
They typically rely on a Transformer-based encoder-decoder architecture driven by learnable queries.
However, existing one-stage models face notable challenges, including a lack of semantic specificity in queries, instance-agnostic query initialization and solely relying on visual features for predicate classification. 
Our VOST-SGG model operates within the one-stage ST-SGG paradigm, but explicitly addresses these issues by incorporating VLM-aided 
dual-source query initialization, and a multi-modal feature bank for improved predicate classification.

\paragraph{LLMs and scene graphs. }
Recent research has increasingly explored the integration of LLMs and VLMs with scene graphs.
One major direction focuses on open-ended \cite{dutta2025open} or weakly supervised \cite{kim2024llm4sgg} scene graph generation, where LLMs are used to infer relational structures without strict supervision or fixed vocabularies. These approaches aim to predict unseen objects and predicates
often by leveraging VLMs through zero-shot \cite{zhao2023less} or instruction-tuning \cite{xu2025llava}. 
Unlike open-ended scene graph generation, we leverage the common-sense knowledge embedded in VLMs to overcome the inherent limitations of existing close-ended one-stage ST-SGG models.


\section{Method}
\label{sec:method}

\subsection{Problem Formulation}




Formally, a scene-graph is composed of a set of relational triplets $\langle s, p, o \rangle$ where \(s\), \(p\), and \(o\) represent the subject, predicate, and object, respectively, drawn from a pre-defined vocabulary of object and predicate categories.
Each subject/object instance is defined by its category label and spatial location (i.e., bounding box), and each edge captures a predicate expressing the relationship between a subject and an object. 
Given a video sequence, our objective is to generate a scene graph for each target frame by modelling the conditional probability of the triplet set given both the current frame and its surrounding frames.

We apply our two key architectural innovations i.e., dual source query initialization and multi-modal feature bank on \cite{oed}, a DETR-style \cite{detr} one-stage formulation that directly predicts relational triplets from the current frame and its reference context, enabling joint spatio-temporal reasoning and efficient end-to-end training.

\subsection{Review of DETR architecture}
\label{sec:detr}

DETR~\cite{detr} formulates object detection as a set prediction problem. A convolutional neural network (CNN)~\cite{he2016deep} is used to extract visual features \( \mathbf{f} \in \mathbb{R}^{C \times H \times W} \) from an image \( \mathbf{I} \in \mathbb{R}^{3 \times H_0 \times W_0} \), where \( H_0, W_0 \) and \( H, W \) denote the input and feature map dimensions, respectively.
Positional encodings \( \mathbf{f}_{\text{pe}} \) are added to these features and are passed to a Transformer encoder. Multi-head self-attention is then applied, computing the query, keys and value as,
\begin{equation}
\resizebox{0.6\columnwidth}{!}{%
$\mathbf{E} = \text{MHead}(\mathbf{Q=f}, \mathbf{K=f}, \mathbf{V=f}). 
\label{eq:encoder}$%
}
\end{equation}

\paragraph{Learnable queries.}
\label{para:learnable_queries}

In standard DETR formulation, each learnable object query \( \mathcal{Q}^{\text{o}} \in \mathbb{R}^{N \times d} \), consists of two components:  
a \emph{content query} \( \mathcal{Q}_{\text{content}} \) and a \emph{position query} \( \mathcal{Q}_{\text{pos}} \). Here, \( N\) is the number of object queries and \(d\) is the hidden dimension.
These components are combined by simple addition as, 
\begin{equation}
    \mathcal{Q}^{\text{o}} = \mathcal{Q}_{\text{content}} + \mathcal{Q}_{\text{pos}}.
    \label{eq:learnable_query}
\end{equation}
This is similar to the typical positional encoding, where positional embeddings are added to image features before feeding into the Transformer encoder. 
Prior works \cite{yao2021efficient, liu2022dab} have suggested that the content query is responsible for specifying what the model should focus on, whereas the position query guides where to attend. We adopt this interpretation in our work. Therefore, the content query \( \mathcal{Q}_{\text{content}} \) guides semantic intent (i.e., “what” to attend), while the position query \( \mathcal{Q}_{\text{pos}} \) provides spatial guidance (i.e., “where” to attend). 

Accordingly, the resulting query \( \mathcal{Q}^{\text{o}} \) is first passed through a multi-head self-attention to enable 
interaction among queries as in Eq.~\ref{eq:decoder_a}. The updated query representation \( \mathcal{Q}^{\text{obj}} \) then attends to 
encoded image features \( \mathbf{E} \in \mathbb{R}^{L \times d} \) via cross-attention, where 
\( L = H \times W \) denotes the flattened encoded image feature map as shown in Eq. \ref{eq:decoder_b},

\begin{subequations} \label{eq:decoder} 
\begin{align}
\resizebox{0.7\columnwidth}{!}{$
\mathbf{\mathcal{Q}^{\text{obj}}} = \text{MHead}(\mathbf{Q=\mathcal{Q}^{\text{o}}}, \mathbf{K=\mathcal{Q}^{\text{o}}}, \mathbf{V=\mathcal{Q}^{\text{o}}});
$} \label{eq:decoder_a} \\[0.3em]
\resizebox{0.7\columnwidth}{!}{$
Q^{\text{obj}'} = \text{MHead}(\mathbf{Q=\mathcal{Q}^{\text{obj}}}, \mathbf{K=\mathbf{E}}, \mathbf{V=\mathbf{E}}).
$} \label{eq:decoder_b}
\end{align}
\end{subequations}

\noindent Finally, the decoder output ($Q^{\text{obj}'}$) is passed to an MLP and a classifier to predict bounding boxes and the object class respectively. A Hungarian algorithm is used to match predictions to ground truth for training.

\subsection{Overview of VOST-SGG}
\label{sec:model-overview}


\begin{figure}[ht]
    \includegraphics[width=\linewidth]{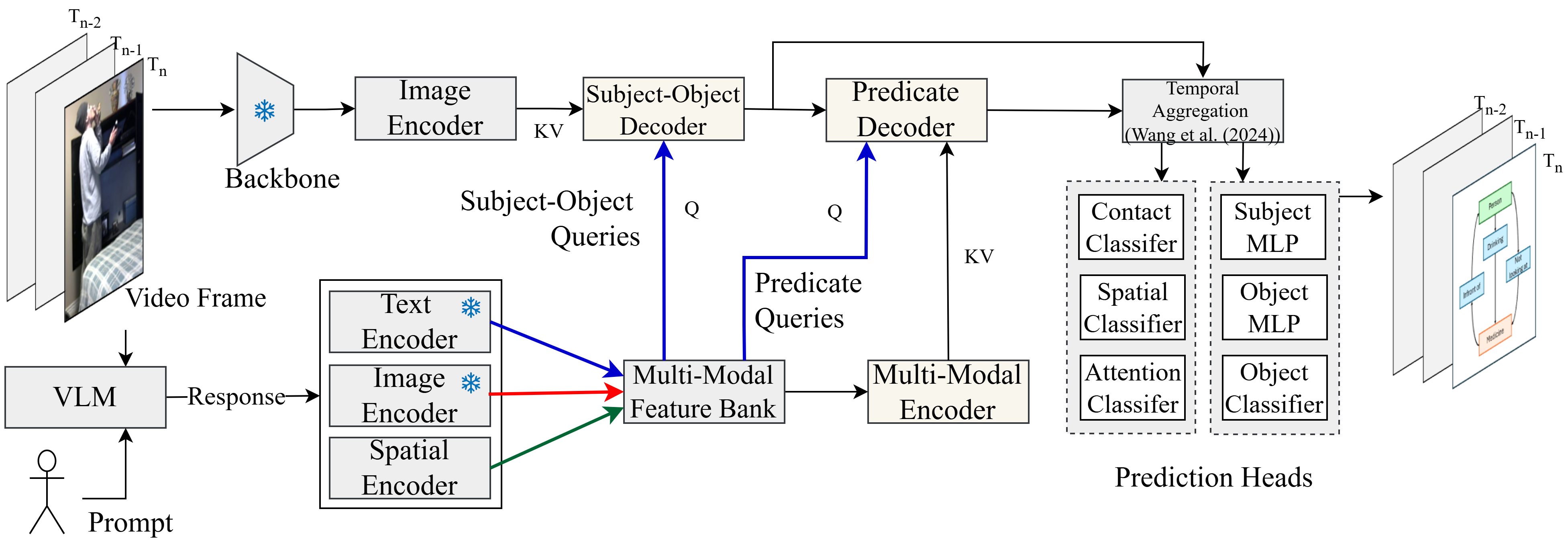}
    \caption{Overall pipeline of our VOST-SGG framework. 
        Subject-object and predicate queries attend to visual and VLM-aided multi-modal features respectively, followed by 
        prediction heads to generate the final spatio-temporal scene graph. \textcolor{blue}{Blue}, \textcolor{red}{Red} and \textcolor{ForestGreen}{Green} lines depict the textual, visual and spatial embeddings.}
    \label{fig:main_architecture}
\end{figure}

\begin{figure}[ht]
\centering
    \includegraphics[width=0.8\linewidth]{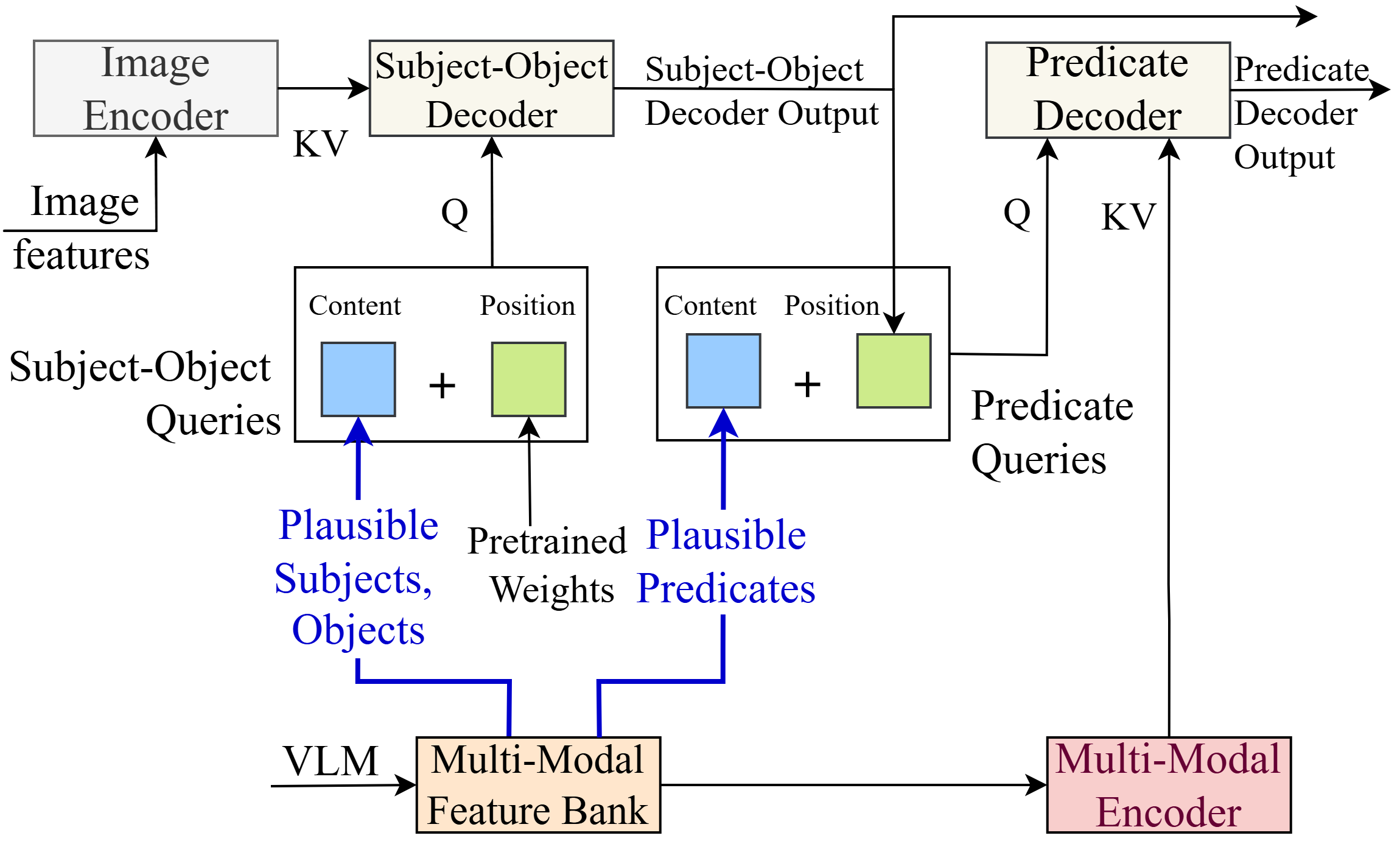}
    \caption{Dual source query initialization process.}
    \label{fig:dual_source_architecture}
\end{figure}

\begin{figure}[ht]
\centering
    \includegraphics[width=0.8\linewidth]{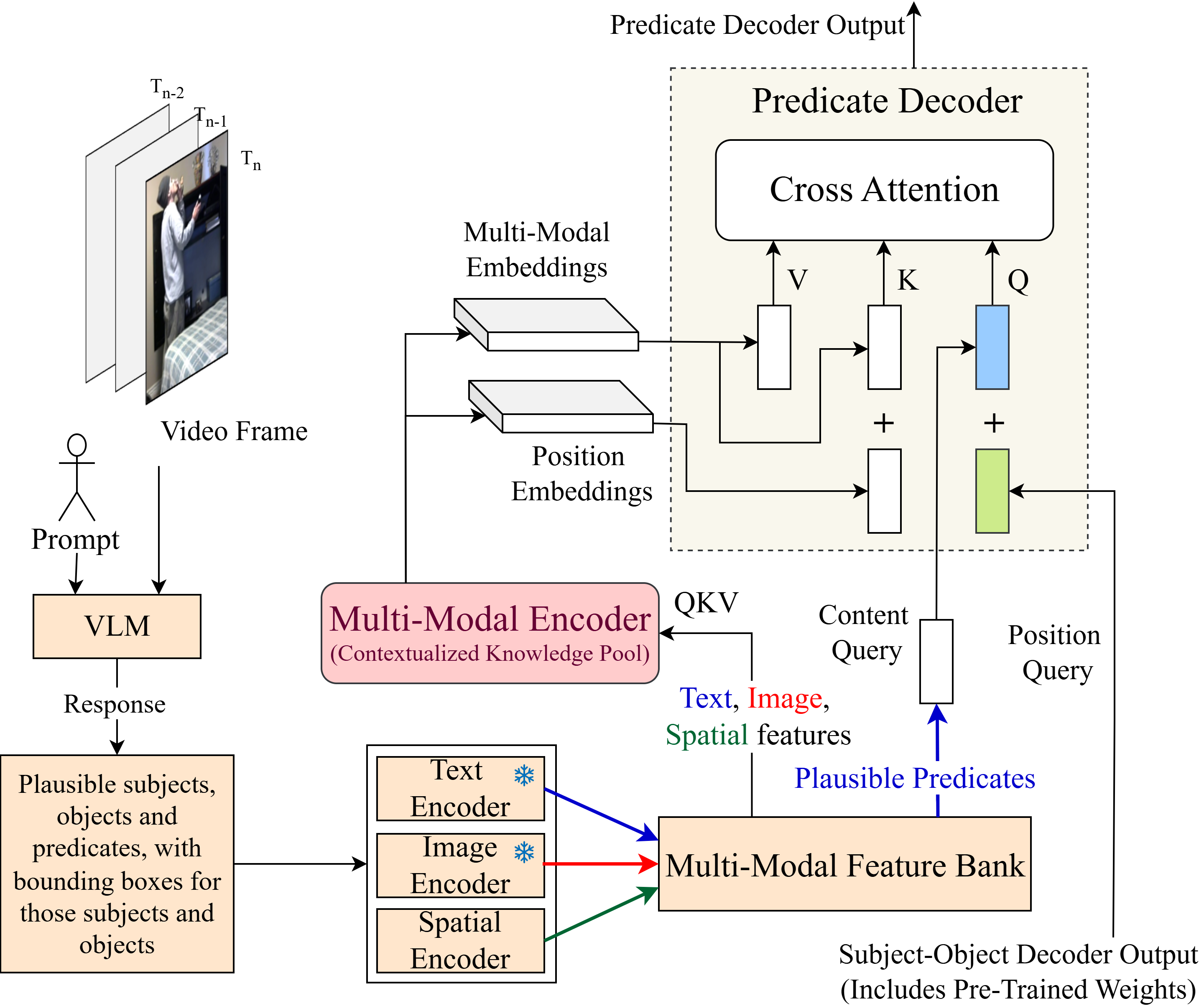}
    \caption{Embeddings derived from multi-modal feature bank, and the predicate decoder.}
    \label{fig:mm_architecture}
\end{figure}

The key architectural innovations of VOST-SGG lie in its query re-design (Fig. \ref{fig:dual_source_architecture} and Sec. \ref{sec:query_redesign}) and the introduction of a multi-modal feature bank for predicate classification (Fig. \ref{fig:mm_architecture} and Sec. \ref{sec:mm}). These components are built on top of the OED framework \cite{oed}, which follows DETR-style encoder-decoder architecture \cite{detr}. 
Internal architectures of the encoders and decoders are shown in Supplementary Section~\ref{app:encoder_architecture}.
%
Our end-to-end pipeline is illustrated in Fig.~\ref{fig:main_architecture}. Given a video frame, we first extract visual features using an image backbone followed by an image encoder. In parallel, we prompt a VLM to generate plausible subject, object, and predicate candidates for the given frame. These VLM responses are then embedded through a series of multi-modal encoders to construct a multi-modal feature bank. These features are further refined via a dedicated multi-modal encoder, while they are also used to guide the formulation of subject-object and predicate queries used in the subsequent subject-object and predicate decoders.
Next, subject-object queries cross-attend with the encoded image features within the subject-object decoder to capture subject-object features. 
Then, predicate queries cross-attend to the multi-modal features within the predicate decoder to extract predicate features. 
The subject–object and predicate queries are further enriched with temporal context during the temporal aggregation stage and are then passed to their respective prediction heads to produce the final spatio-temporal scene graph.

\paragraph{Image and multi-modal encoders.}
Image encoder takes backbone features and outputs encoded features of the target frame $T$ and $n$ reference frames  \( \mathcal{F} = \{ \mathbf{F}_T, \mathbf{F}_{T_1}, \dots, \mathbf{F}_{T_n} \} \) as in Eq. \ref{eq:encoder}.
We also prompt VLM to generate plausible subject, object, and predicate candidates for each frame. These VLM responses are then embedded through a series of multi-modal encoders to construct multi-modal feature bank \(\mathcal{F}^{MMBank}\) (for details refer Eq.~\ref{eq:mm_bank} and Sec.~\ref{sec:mm}). Multi-modal encoder takes these features and output encoded multi-modal features \( \mathcal{F}^{MM} = \{ \mathbf{F}^{MM}_T, \mathbf{F}^{MM}_{T_1}, \dots, \mathbf{F}^{MM}_{T_n} \} \).
Both \( \mathbf{F}_i \) and \( \mathbf{F}^{MM}_i \) are augmented with positional encodings as shown in 
Supplementary~\ref{app:encoder_architecture} Fig. \ref{fig:encoder_internal}.

\paragraph{Cascaded scene graph decoders.}
To effectively model the scene graph generation task, we use two cascaded Transformer decoders: (1) subject-object decoder and (2) predicate decoder as shown in Fig. \ref{fig:main_architecture}. These modules are designed to sequentially model the entity-level $s,o$ and relation-level $p$ components of the relational triplets respectively. Both decoders utilize multi-head attention mechanisms, as described in Eq. \ref{eq:decoder} and as illustrated in Supplementary~\ref{app:encoder_architecture} Fig. \ref{fig:decoder_internal}.

\paragraph{Subject-object decoder based on subject-object queries.}
In contrast to DETR, which relies on object-centric queries, our framework requires queries that explicitly encode elements in the relational triplets. To this end, we define a set of learnable subject-object queries \( \mathcal{Q}^{s-o} \in \mathbb{R}^{N \times d} \), where each query initially corresponds to a candidate subject-object pair \( (s, o) \).
The Subject-object decoder operates on the subject-object queries \( \mathcal{Q}^{s-o} \), where the content query is initialized with embeddings of plausible subject-object pairs retrieved from VLM and the position query is initialized using pre-trained weights as in Fig. \ref{fig:dual_source_architecture} (details in Sec.~\ref{sec:query_redesign}). These queries are processed through self-attention and cross-attention with the target frame features \( \mathbf{F}_T \) as given in Eq. \ref{eq:decoder}. This yields subject-object aware queries \( \mathcal{Q}^{\text{s-o}'} = \{ q_1^{s-o'}, \ldots, q_{N}^{s-o'} \} \in \mathbb{R}^{N \times d} \) representing each subject-object pair.

\paragraph{Predicate decoder based on predicate queries.}
This is responsible for modelling predicate features corresponding to each subject-object pair. Predicate queries \( \mathcal{Q}^{p} \) are constructed using plausible predicates retrieved from the VLM, which serve as the content queries, while \( \mathcal{Q}^{s-o'} \) act as position queries (details in Sec.~\ref{sec:query_redesign}). These predicate queries then cross-attend with the multi-modal embeddings \( \mathcal{F}^{\text{MM}} \) as in Eq.~\ref{eq:mm}, resulting in predicate-aware queries \( \mathcal{Q}^{p'} = \{ q_1^{p'}, \ldots, q_{N}^{p'} \} \in \mathbb{R}^{N \times d} \).

\begin{equation}
\resizebox{0.8\columnwidth}{!}{
$\mathcal{Q}^{p'} = \text{MHead}(\mathbf{Q} = \mathcal{Q}^{p}, \mathbf{K} = \mathcal{F}^{\text{MM}}, \mathbf{V} = \mathcal{F}^{\text{MM}}).$
}
\label{eq:mm}
\end{equation}


\paragraph{Relational triplet representation.}
We concatenate both subject-object and predicate queries to form the relational triplet representation, 

\begin{equation}
\mathcal{Q}^{tri} = \text{Concat}(\mathcal{Q}^{s-o'}, \mathcal{Q}^{p'}) = \{ q_1^{tri}, \ldots, q_{N}^{tri} \} \in \mathbb{R}^{N \times 2d}.
\label{eq:concat}
\end{equation}

\noindent Each embedding \( q^{tri}_i \) represents a candidate triplet \( \langle s, p, o \rangle \) where \( q^{tri}_i = Concat(q_i^{s-o}, q_i^{p})\) and \( i \in \{1, \dots, N\}\). 
%
The triplet representations of the current frame and $n$ reference frames \(\{ \mathcal{Q}_T^{tri}, \mathcal{Q}_{T_1}^{tri}, \dots, \mathcal{Q}_{T_n}^{tri} \}\), are fed into a temporal aggregation module to incorporate context from surrounding frames into the current frame \(\mathcal{Q}_{T}\). We use the temporal aggregation module proposed by \citet{oed}. 

\paragraph{Prediction heads.}
The resulting temporally enriched representation of the current frame denoted by \(\mathcal{Q}_{T}^{tri'}\) is then split into: (1) subject-object level features \( \mathcal{Q}^{s-o'}_T \in \mathbb{R}^{N \times d} \) and (2)  predicate-level features \( \mathcal{Q}^{p'}_T \in \mathbb{R}^{N \times d} \).
The subject-object features \( \mathcal{Q}^{s\text{-}o'}_T \) are passed through subject and object detection MLP heads and classifiers, to predict the bounding boxes and class labels respectively for the subject and object entities in each triplet.
The predicate-level features \( \mathcal{Q}^{p'}_T \) are forwarded to \(\mathfrak{p}\) number of predicate classifiers which are responsible for classifying predicates.

    
    


\subsection{Re-visiting query design: A dual-source approach for "what-where" reasoning}
\label{sec:query_redesign}

As discussed in Sec. \ref{sec:detr}, each learnable query in DETR-like models consists of two components: (1) content query guides what the model should attend to and (2) position query guides where in the image the model should focus \cite{liu2022dab}. Most existing work \cite{detr, wang2022anchor,zhu2010deformable} including recent methods \cite{oed}, primarily emphasize only the spatial aspect focusing on where to look. In these setups, although the content query is intended to encode what to attend to, it is typically assigned with a zero vector, providing no semantic guidance. Meanwhile, the position query is initialized either randomly or with pre-trained weights \cite{detr, oed} or using coarse spatial anchors \cite{wang2022anchor, liu2022dab} (see Fig. \ref{fig:intro_diagram_oed}). This setup has several core limitations: \textbf{(1) Lack of semantic specificity:} queries provide no intent or indication of the object or predicate features they are meant to retrieve and, \textbf{(2) Instance-agnostic initialization:} queries ignore instance-specific information from individual video frames during initialization, limiting their discriminative power as shown in Fig. \ref{fig:intro_diagram_oed}. Here, we use the term \emph{instance} to refer to a single video frame.

\paragraph{Shifting from "where" to "what-where" with dual source query initialization.}
We address the above two limitations using dual source query initialization strategy that shifts the inductive bias of query design, from purely spatial reasoning ("where") to a semantically grounded spatial reasoning approach that combines both "what" and "where" as shown in Fig. \ref{fig:dual_source_architecture}. 
Rather than treating each query as a generic spatial probe, we enrich the content query with semantic cues that define "what" to look for, while the position query continues to guide "where" to look.
Specifically, we enrich the content query with an instance-specific semantic prior i.e., plausible object names and predicate names, derived from common sense knowledge of a VLM. These high-level concepts (e.g., “person”, “hold”, “cup”) give each query an explicit head start on the type of relational triplet to retrieve, acting as \emph{weak, frame-conditioned priors that constrain the search space and facilitate further refinement by the model}.
In contrast, the position queries are instance-agnostically initialized using pre-trained weights \(\mathcal{W} \in \mathbb{R}^{N \times d}\) from MS-COCO and later fine-tuned on Action Genome \cite{tpt}, providing spatial anchors that offer spatial stability while retaining flexibility across varying scenes.
In this way, our dual-source query initialization strategy assigns distinct roles to the content and position components of each learnable query based on instance-specificity. 

We pass each video frame to a VLM and prompt for the plausible subjects-object pairs, and their predicates and obtain the text embeddings of subject, object and predicate using text encoder $\psi_t$ resulting in \( sub_t, obj_t, pred_t \in \mathbb{R}^{d_l} \). \( d_l \) denotes the language embedding dimension.
In subject-object decoder, we assign the concatenation of $sub_t$ and $obj_t$ embeddings as \(\text{Concat}(sub_t, obj_t) \in \mathbb{R}^{2d_l}\) into $x$ out of $N$ number of content queries in $\mathcal{Q}$ as in Eq. \ref{eq:sub-obj_query}. $x$ is the number of plausible subject-object pairs prompted by the VLM for the given frame.

\begin{equation}
\resizebox{0.9\columnwidth}{!}{%
$\begin{aligned}
    \mathcal{Q}^{s-o}_{\text{content}} &= \text{Linear}(\text{Concat}(sub_t, obj_t)) \in \mathbb{R}^{N \times d}; \quad
    \mathcal{Q}^{s-o}_{\text{pos}} &= \mathcal{W}; \\
    \mathcal{Q}^{s-o} &= \mathcal{Q}^{s-o}_{\text{content}} + \mathcal{Q}^{s-o}_{\text{pos}} \in \mathbb{R}^{N \times d}.
\end{aligned}$%
}
\label{eq:sub-obj_query}
\end{equation}

\noindent In the predicate decoder, we assign the relevant predicate embeddings $pred_t$ to the content query, while learned output subject-object queries \(\mathcal{Q}^{s-o'}\) serve as the position query to form the predicate queries \(\mathcal{Q}^{p}\) as, 

\begin{equation}
\resizebox{0.8\columnwidth}{!}{%
$\begin{aligned}
    \mathcal{Q}^{p}_{\text{content}} &= \text{Linear}(pred_t) \in \mathbb{R}^{N \times d}; \quad
    \mathcal{Q}^{p}_{\text{pos}} &= \mathcal{Q}^{s-o'};  \\
    \mathcal{Q}^{p} &= \mathcal{Q}^{p}_{\text{content}} + \mathcal{Q}^{p}_{\text{pos}} \in \mathbb{R}^{N \times d}.
\end{aligned}$%
}
\label{eq:predicate_query}
\end{equation}


\subsection{Multi-modal feature bank}
\label{sec:mm}

While recent one-stage ST-SGG models \cite{oed} have advanced ST-SGG by enabling end-to-end training, they are predominantly \textit{uni-modal}, relying solely on visual features. This uni-modal design restricts the model’s ability to reason about complex interactions that go beyond appearance, especially in semantically rich scenarios which may look visually similar but differ in functional context, e.g., distinguishing between “person opening a door” and “person closing a door” requires certain contextual knowledge to disambiguate varied semantics of the scene. 

To overcome this limitation, we propose the use of multi-modal encoder shown in Fig. \ref{fig:mm_architecture}. This encoder aggregates both visual and common sense knowledge. 
Apart from the plausible subject-object pairs, and their predicates in a frame, we also prompt the VLM for the bounding boxes of the subjects and objects. Based on the bounding box responses, we extract the region-specific visual embeddings for the subject, object, and predicate using visual encoder $\psi_v$ as $sub_v$ $obj_v$ and $pred_v$ respectively.
We also convert the VLM-predicted bounding boxes into a single spatial feature and  project it into the same space as the visual and textual features, obtaining a unified spatial embedding for the relationship region $i$ as,



\begin{equation}
    s_i = 
    \text{Linear}(\text{Concat} ( 
        \mathrm{b}^{sub}_i , 
        \mathrm{b}^{obj}_i , 
        \big( \mathrm{c}^{sub}_i - \mathrm{c}^{obj}_i \big) , 
        \mathrm{a}^{sub}_i ,
        \mathrm{a}^{obj}_i)),
    \label{eq:spatial_feature}
\end{equation}

\noindent where $s_i \in \mathbb{R}^d$, $\text{Concat}(\cdot) \in \mathbb{R}^{12}$, \(\mathrm{b, c, a}\) refer to bounding box coordinates, center coordinates and area of subject and object bounding boxes respectively.
This compact spatial representation captures the relative position and scale between the subject and object, enabling the model to better localize potential regions of interaction \cite{oed}. 
Accordingly, we form the multi-modal feature bank $\mathcal{F}^{\text{MMBank}}$ with visual, textual and spatial features
as in Eq. \ref{eq:mm_bank}, followed by multi-head self attention in multi-modal encoder forming the multi-modal embeddings $ \mathcal{F}^{\text{MM}}$ as in Eq. \ref{eq:encoder}.



\begin{equation}
    \mathcal{F}^{\text{MMBank}} = \big\{ sub_{m_i}, obj_{m_i}, pred_{m_i}, s_i \big\}_{i=1}^{x}.
    \label{eq:mm_bank}
\end{equation}

\noindent Here, \( m_i \in \{v_i, t_i\} \) indicates visual and textual modalities associated with the $i^{th}$ region.
The proposed multi-modal encoder produces frame-specific contextual representations grounded in textual, visual, and spatial common-sense evidence, enabling robust predicate classification. Under our dual-source query initialization, \emph{content} queries carry instance-specific 'what' semantics from VLM responses, while \emph{position} queries are initialized instance-agnostically with pre-trained weights. This gives each query a semantically grounded head start (e.g., a sofa in front of a person), guiding it toward plausible concepts rather than exploring the feature space blindly with zeros. 
However, it should be noted that, in dual-source query initialization VLM outputs serve only as initialization: during training, the model learns to refine and, when necessary, correct these hints using the frame-specific context provided by the multi-modal feature bank. When VLM cues are noisy or inaccurate, the multi-modal feature bank supplies complementary textual, visual, and spatial signals from current and reference frames providing alternative gradient pathways via shared projections, so the model can focus on the most informative and useful evidence. This design is particularly beneficial for rare predicates in long-tailed distributions, resulting in more balanced scene graph generation.

\paragraph{Training and optimization}

Given a video sequence, the model generates a fixed set of predictions for each frame. The Hungarian algorithm finds the optimal one-to-one matching 
\( \hat{\sigma} \in \mathfrak{S}\) 
between prediction set \( P = \{p_i\}_{i=1}^{N}\) and ground truth \( G = \{g_i\}_{i=1}^{N}\) which is padded with \(\emptyset\) to maintain a consistent length.

\begin{equation}
\resizebox{0.7\columnwidth}{!}{%
$
\hat{\sigma} = \arg\min_{\sigma \in \mathfrak{S}_N} \sum_{i}^{N} \mathcal{L}_{\text{match}}\left(g^i, p^{\sigma(i)}\right),
\label{eq:hungarian}$%
}
\end{equation}

\noindent where $\mathcal{L}_{\text{match}}(g_i, p^{\sigma(i)})$ denotes the pairwise matching cost between ground truth \( g_i \) and the prediction at index \( \sigma(i) \).
The matching loss is defined as,

\begin{equation}
\resizebox{0.9\columnwidth}{!}{%
$
\mathcal{L}_{\text{match}}\left(g^i, p^{\sigma(i)}\right) = \sum_{j \in \{s, o, p\}} \alpha_j \mathcal{L}^{j}_{\text{cls}} + \beta \sum_{j \in \{s, o\}} \mathcal{L}^{j}_{\text{box}}
\label{eq:match_loss}$%
}
\end{equation}

\noindent where $\mathcal{L}^{j}_{\text{cls}} = \mathcal{L}_{\text{cls}}^{j}\left(g_j^i, p_j^{\sigma(i)}\right),\ j \in \{s, o, p\}$ indicates the classification loss for subject, object and predicate, and 
$\mathcal{L}^{j}_{\text{box}} = \mathcal{L}_{\text{box}}^{j}\left(g_j^i, p_j^{\sigma(i)}\right),\ j \in \{s, o\}$ indicates the bounding box regression loss. Following \cite{oed}, we use cross-entropy loss as the classification loss of subject $\mathcal{L}^{s}_{\text{cls}}$ and object $\mathcal{L}^{o}_{\text{cls}}$, a weighted sum of $L_1$ loss and GIoU loss~\cite{rezatofighi2019generalized} as the bounding box regression loss of subject $\mathcal{L}^{s}_{\text{box}}$ and object $\mathcal{L}^{o}_{\text{box}}$, and focal loss~\cite{lin2017focal} as the classification loss of predicate.

\section{Experiments}
\label{sec:experiments}

\subsection{Experimental settings}

\paragraph{Datasets. }
We evaluate our model on the Action Genome \cite{ag}, which extends the Charades by providing detailed ST-SGG annotations for 234K sampled frames across approximately 10K videos. Each annotation includes object instances and their relationships, spanning 35 object categories and 25 predicate types. The predicates are grouped into \(\mathfrak{p}=3\) categories: attention, spatial, and contacting.

\paragraph{Evaluation metrics. }
We evaluate VOST-SGG using the standard scene graph evaluation metrics Recall@K (R@K) for $\text{K} \in \{10, 20, 50\}$. We also report mean Recall@K (mR@K) \cite{tempura}.
We evaluate our model on two settings: (1) Scene Graph Detection (SGDET), which requires detecting subject-object pairs along with their corresponding predicates
and, (2) predicate Classification (predCLS), where, ground truth bounding boxes are provided for subject-object pairs, and the task is to classify the correct predicate. 
We report performance under two constraint settings: With and No Constraints, where in the former, the subject-object pairs are restricted to have at most one predicate and the latter can have multiple predicates. 

\paragraph{Implementation details. }
We use the Qwen2.5-VL-7B-Instruct model~\cite{bai2025qwen2} as VLM due to its strong visual-semantic capabilities. We instruct-tune Qwen on the Action Genome dataset~\cite{ag} to better align it with ST-SGG objectives. Example prompts used for the PredCLS and SGDET settings are given in 
Supplementary \ref{app:prompts}.
We use ResNet-50 as the image backbone and Clip ViT B32 text and image encoders for extracting textual and visual embeddings of VLM object and predicate cues. When extracting region-specific visual features, we use the translucent background prompt \cite{sugandhika2025situational}. 
All encoders and decoders use 6 transformer layers, with the number of learnable queries $N = 100$ \cite{oed, detr}. 
We train and evaluate our models on a single node equipped with 8 NVIDIA RTX A6000 GPUs, each with 49GB of memory.

\subsection{Main results}
\label{subsec:main_results}

We report the performance of our proposed method, VOST-SGG, on the Action Genome in Tab.~\ref{tab:main_results}, comparing it against:
(1) one-stage ST-SGG model OED; (2) multi-stage ST-SGG models: TPT, DSG-DETR, TEMPURA, TR2, STTran-TPI, APT, STTran, and TRACE; and (3) image-based scene graph generation models evaluated on Action Genome: GPS-Net, RelDN, VCTree, M-FREQ, MSDN and VRD.
One-stage models integrate object detection, temporal context modeling, and predicate classification into a single end-to-end framework, whereas multi-stage models perform these steps separately.
The results offer several key observations.
First, one-stage models tend to outperform their multi-stage counterparts, highlighting the benefits of joint end-to-end optimization and cross-task synergy.
Second, multi-stage ST-SGG models consistently outperform image-based methods, largely due to their ability to exploit temporal context. 
Finally, our method, VOST-SGG, surpasses the previous state-of-the-art OED by a notable margin in both Recall and mean Recall, demonstrating the strengths of our proposed LLM-guided query re-design and multi-modal feature bank. 


\begin{table*}[ht]
\centering
\resizebox{1.0\linewidth}{!}{
\begin{tabular}{|l|c|c|c|c|c|c|c|c|}
\toprule

\multirow{3}{*}{Method} & \multicolumn{4}{c|}{predCLS} &  \multicolumn{4}{c|}{SGDET} \\
\cline{2-9}

& \multicolumn{2}{c|}{With Constraint} & \multicolumn{2}{c|}{No Constraint} & \multicolumn{2}{c|}{With Constraint} & \multicolumn{2}{c|}{No Constraint} \\
\cline{2-9}

&   R@10 / 20 / 50 & mR@10 / 20 / 50 & R@10 / 20 / 50 & mR@10 / 20 / 50 & R@10 / 20 / 50 & mR@10 / 20 / 50 & R@10 / 20 / 50 & mR@10 / 20 / 50 \\
\midrule
VRD \cite{lu2016visual}         & 51.7 / 54.7 / 54.7 &   -   & 59.6 / 78.5 / 99.2    & -   & 19.2 / 24.5 / 26.0   &  -   & 19.1 / 28.8 / 40.5     & -   \\
MSDN  \cite{li2017scene}        & 65.5 / 68.5 / 68.5 &   -   & 74.9 / 92.7 / 99.0    & -   & 24.1 / 32.4 / 34.5 &  -   & 23.1 / 34.7 / 46.5     & -   \\
M-FREQ \cite{zellers2018neural} & 62.4 / 65.1 / 65.1 &   -   & 73.4 / 92.4 / 99.6    & -   & 23.7 / 31.4 / 33.3 &  -   & 22.8 / 34.3 / 46.4     & -   \\
VCTree \cite{tang2019learning}  & 66.0 / 69.3 / 69.3 &   -   & 75.5 / 92.9 / 99.3    & -   & 24.4 / 32.6 / 34.7 &   -  & 23.9 / 35.3 / 46.8     & -   \\
RelDN  \cite{zhang2019graphical}& 66.3 / 69.5 / 69.5 & 6.2 / 6.2 / 6.2 & 75.7 / 93.0 / 99.0  & 31.2 / 63.1 / 75.5 & 24.5 / 32.8 / 34.9 & 3.3 / 3.3 / 3.3 & 24.1 / 35.4 / 46.8 & 7.5 / 7.5 / 7.5 \\
GPS-Net \cite{lin2020gps}       & 66.8 / 69.9 / 69.9 &   -   & 76.0 / 93.6 / 99.5   & -    & 24.7 / 33.1 / 35.1 &   -  & 24.4 / 35.7 / 47.3     & -   \\

\midrule
TRACE  \cite{trace}  & 27.5 / 27.5 / 27.5 & 15.2 / 15.2 / 15.2 & 72.6 / 91.6 / 96.4 & 50.9 / 73.6 / 82.7 & 13.9 / 14.5 / 14.5 & 8.2 / 8.2 / 8.2 & 26.5 / 35.6 / 45.3 & 22.8 / 31.3 / 41.8 \\
STTran \cite{sttran} & 68.6 / 71.8 / 71.8 & 37.8 / 40.1 / \underline{40.2} & 77.9 / 94.2 / 99.1 & 51.4 / 67.7 / 82.7 & 25.2 / 34.1 / 37.0   & 20.8 / 20.8 / 22.2 & 24.6 / 36.2 / 48.8 & 20.9 / 29.7 / 39.2 \\
APT \cite{apt}       & 69.4 / 73.8 / 73.8 & -   & 78.5 / 95.1 / 99.2 & -   & 26.3 / 36.1 / 38.3 & -   & 25.7 / 37.9 / 50.1 & -   \\
STTran-TPI \cite{wang2022dynamic}   & 69.7 / 72.6 / 72.6 & 37.3 / 40.6 / 40.6 & -   & -   & 26.2 / 34.6 / 37.4 & 20.2 / 20.2 / 22.8 & -   & -   \\
TR2 \cite{tr2}       & 70.9 / 73.8 / 73.8 & -   & 83.1 / \underline{96.6} / \textbf{99.9} & -   & 26.8 / 35.5 / 38.3 & -   & 27.8 / 39.2 / 50.0   & -   \\
TEMPURA \cite{tempura}  & 68.8 / 71.5 / 71.5 & \underline{42.9} / \underline{46.3} / \underline{40.2} & 80.4 / 94.2 / 99.4 & \underline{61.5} / \underline{85.1} / \textbf{98.0} & 28.1 / 33.4 / 34.9 & \underline{22.6} / 22.6 / 23.7 & 29.8 / 38.1 / 46.4 & -   \\
DSG-DETR \cite{feng2023exploiting}     & -   & -   & -   & -   & 30.3 / 34.8 / 36.1 & -   & 32.1 / 40.9 / 48.3 & 24.7 / 33.9 / 43.7 \\
TPT \cite{tpt} & -   & -   & -   & -   & -   & -   & 32.0 / 39.6 / 51.5 & -   \\
DIFFVSGG \cite{chen2025diffvsgg}   & 71.9 / 74.5 / 74.5 & -   & 83.1 / 94.5 / 99.1 & -   & 32.8 / 39.9 / 45.5 & -   & \underline{35.4} / 42.5 / 51.0   & -   \\
\midrule

OED \cite{oed}  & \underline{73.0} / \underline{76.1} / \underline{76.1} & 42.1 / 46.1 / \underline{46.1} & \underline{83.3} / 95.3 / 99.2 & 56.4 / 84.0 / 94.6 & \underline{33.5} / \underline{40.9} / \underline{48.9} & 20.8 / \underline{26.5} / \underline{32.5} & 35.3 / \underline{44.0} / \underline{51.8} & \textbf{32.6} / \underline{39.1} / \underline{49.7} \\

VOST-SGG (Ours) & \textbf{76.1} / \textbf{79.5} / \textbf{79.5} & \textbf{46.5} / \textbf{50.9} / \textbf{50.9} & \textbf{85.7} / \textbf{97.0} / \underline{99.8} & \textbf{63.3} / \textbf{85.5} / \underline{95.3} & \textbf{34.2} / \textbf{41.5} / \textbf{49.0} & \textbf{23.0} / \textbf{29.7} / \textbf{35.8} & \textbf{36.5} / \textbf{44.8} / \textbf{52.6} & \underline{28.7} / \textbf{41.6} / \textbf{51.5} \\

\bottomrule
\end{tabular}
}
\caption{Comparison with state-of-the-art spatio-temporal scene graph generation methods on Action Genome dataset \cite{ag}. The bold and underline font shows the best and the second best result, respectively.}
\label{tab:main_results}
\end{table*}

\begin{figure}[ht]
    \centering
    \includegraphics[width=\linewidth]{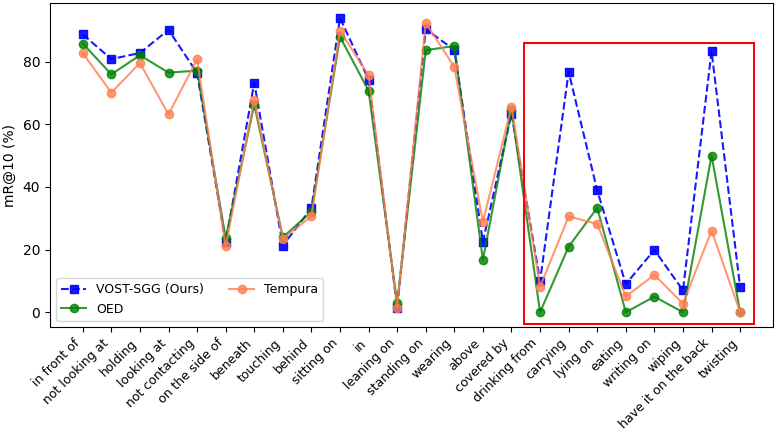}
    \caption{Per-predicate mR@10 (PredCLS, with constraint). We compare VOST-SGG to the single-stage OED~\cite{oed} and multi-stage TEMPURA~\cite{tempura}. Predicates are sorted in descending order by frequency from left to right.}
    \label{fig:mR_line_predcls}
\end{figure}

Notably, as illustrated in Fig.\ref{fig:mR_line_predcls} our model improves performance on rare (tail) predicates such as \emph{wiping}, \emph{writing on}, \emph{have it on the back}, and \emph{twisting}, without sacrificing performance in frequent classes. Similar graph is provided for SGDET setting in 
Supplementary \ref{app:graphs}.
We define “rare” predicates as the rightmost eight predicates in Fig.~\ref{fig:mR_line_predcls} (highlighted by the red rectangle). To analyze long-tail robustness, we compared model-correct predictions with VLM responses for both rare and non-rare categories, using the multi-modal feature bank and, for comparison, conventional image features.
Under the multi-modal feature bank setting, 18\% of model-correct cases for rare predicates overlapped with VLM responses, while the model successfully corrected 16\% of incorrect VLM cues. In contrast, for non-rare predicates, 70\% of model-correct cases aligned with VLM cues, with the model recovering 3\% of incorrect cues.
When the multi-modal feature bank is replaced with conventional image features (as in~\cite{oed}) while keeping dual-source query initialization fixed, rare-predicate recovery drops sharply from 16\% to 6\% (and from 3\% to 2\% for non-rare predicates). These trends highlight the roles of the two main components:
(1) Dual-source query initialization: 
Provides a semantically informed warm start, which benefits non-rare predicates where VLM signals are trustworthy, while still allowing the model to refine or correct noisy VLM cues for rare predicates
(2) Multi-modal feature bank: 
Acts as a frame-specific contextual memory that fuses textual priors, region-level visual evidence, and spatial configurations, enabling the model to override unreliable VLM cues, reflected in the substantial gain from 6\% to 16\% recovery for rare predicates.
When VLM cues for a given frame are unreliable, the multi-modal feature bank further supplies complementary visual and spatial evidence from neighbouring frames. Further, the decoder attends to the full image feature map from the visual backbone, allowing it to re-localize and correct erroneous VLM guidance through iterative cross-attention, consistent with prior work~\cite{liu2022dab, zhu2010deformable}. 
Together, these mechanisms explain how our model effectively corrects initial VLM cues and, 
why our method achieves stronger long-tail performance than OED and TEMPURA (Fig.~\ref{fig:mR_line_predcls}) without any specific class-balancing strategies.

\paragraph{Qualitative results.}
As shown in Supplementary \ref{app:qualtative_results} Fig.~\ref{fig:qualitative_results_main} and \ref{fig:qualitative_results_main2} highlight that, while both VOST-SGG and OED are good at identifying common predicates, VOST-SGG extends its capability in recognizing rare predicates such as \emph{wiping} and \emph{have it on the back} whereas OED incorrectly classifies it as a more frequent predicate \emph{holding}. As illustrated in 
Supplementary~\ref{app:qualtative_results} Fig.~\ref{fig:missed_obj}, VOST-SGG also excels in challenging visual scenarios, by accurately detecting objects like \emph{laptop}, which OED misses, probably due to the blending of the dark color of the laptop with the table. 
Despite its overall performance, VOST-SGG struggles with fine-grained distinctions, such as differentiating between \emph{looking at} and \emph{not looking at}, which often depend on subtle eye movements. 


\paragraph{Ablation Study.}

We conduct an ablation study on the Action Genome dataset to evaluate the contribution of each component in our framework.
We compare three model variants: (A) Instance-Agnostic Query Initialization (IAQI) i.e., OED, (B) Dual-Source Query Initialization (DSQI), and (C) DSQI + Multi-Modal Feature Bank (MMFB) i.e., our full VOST-SGG model.
In (A), both content and position queries are initialized instance-agnostically (zero vectors for content and pre-trained weights for position).
In (B), we inject instance-specific plausible subject, object, and predicate embeddings into the content queries while keeping the position queries instance-agnostic, demonstrating the benefit of our dual-source initialization. Both (A) and (B) rely on visual-only features for predicate classification.
In (C), we further incorporate our multi-modal features, highlighting its advantage for predicate classification.
The results for predCLS setting are shown in Tab. \ref{tab:main_ablation}, from which we can draw several conclusions.
First, the performance gains of (B) over (A) validates that injecting instance-specific, common-sense priors into learnable queries improves overall performance, allowing the model to focus on what to attend to rather than blindly exploring spatial locations. 
Second, adding the multi-modal features (C) yields an additional performance boost, showcasing the advantage of combining textual and spatial features with visual ones for better predicate disambiguation, especially in rare or ambiguous interactions where visual-only cues are insufficient. 

\begin{table}[ht]
\centering
\resizebox{\columnwidth}{!}{
\begin{tabular}{|l|c|c|c|c|}
\toprule

\multirow{2}{*}{Method} & \multicolumn{2}{c|}{With Constraint} & \multicolumn{2}{c|}{No Constraint}   \\
\cline{2-5}
&   R@10 / 20 / 50 & mR@10 / 20 / 50 & R@10 / 20 / 50 & mR@10 / 20 / 50 \\

\toprule


(A) IAQI  & 73.0 / 76.1 / 76.1 & 42.1 / 46.1 / 46.1 & 83.3 / 95.3 / 99.2 & 56.4 / 84.0 / 94.6 \\

(B) DSQI & \underline{74.7} / \underline{77.9} / \underline{78.0}	& \underline{46.2} / \underline{50.8} / \underline{50.8}	& \underline{84.7} / \underline{96.4} / \underline{99.6}	& \underline{59.6} / \textbf{86.5} / \textbf{98.6} \\


(C) DSQI + MMFB \textbf{(Ours)} & \textbf{76.1} / \textbf{79.5} / \textbf{79.5} & \textbf{46.5} / \textbf{50.9}  / \textbf{50.9} & \textbf{85.7} / \textbf{97.0} / \textbf{99.8} & \textbf{63.3} / \underline{85.5} / \underline{95.3} \\

\bottomrule
\end{tabular}
}
\caption{Ablation study on Action Genome dataset \cite{ag} for query re-design and multi-modal feature bank for predCLS setting. IAQI: Instance-Agnostic Query Initialization, DSQI: Dual-Source Query Initialization, MMFB: Multi-Modal Feature Bank. 
Ablation study for SGDET setting is provided in 
Supplementary \ref{app:ablation} Tab. \ref{tab:app:ablation}.}
\label{tab:main_ablation}
\end{table}

\paragraph{Performance of VLMs on spatio-temporal scene graph generation.}

We evaluate two VLMs Qwen \cite{bai2025qwen2} and InternVL \cite{chen2024internvl} on the ST-SGG task under the predCLS (with constraint) setting, and summarize the results in Tab.~\ref{tab:vlm_performance}, highlighting several key observations. 
First, the noticeably lower performance in the zero-shot setting for both VLMs in (A) and (B) indicates that foundation models alone struggle to effectively address complex vision-language tasks such as ST-SGG. However, when fine-tuned as in (C) and (D), they achieve competitive performance comparable to the state-of-the-art task-sepcific OED model shown in (E).
Furthermore, when VLM knowledge is properly integrated for task-specific models as in (E)–(H), the overall performance improves, demonstrating the benefits of properly integrating pretrained vision-language knowledge with specialized task-driven designs for complex structured vision-language tasks such as ST-SGG.


\begin{table}[ht]
\centering
\resizebox{\columnwidth}{!}{
\begin{tabular}{|l|c|c|c|c|}
\toprule

\multirow{2}{*}{Method} & \multicolumn{2}{c|}{With Constraint} & \multicolumn{2}{c|}{No Constraint}   \\
\cline{2-5}
&   R@10 / 20 / 50 & mR@10 / 20 / 50 & R@10 / 20 / 50 & mR@10 / 20 / 50 \\

\toprule

(A) QWEN Zero-shot \cite{bai2025qwen2} & 38.8 / 39.7 / 39.7 & 25.8 / 26.9 / 27.0 & 41.0 / 48.0 / 78.2 & 27.9 / 39.3 / 69.3 \\
(B) InternVL Zero-shot \cite{chen2024internvl}	& 17.8 / 18.0 / 18.0 & 13.2 / 14.1 / 14.1 & 24.5 / 34.0 / 75.4 & 18.5 / 32.5 / 67.2 \\
\hline
(C) QWEN Fine-tuned \cite{bai2025qwen2}   & 71.7 / 75.6 / 75.6 & 56.2 / 57.8 / 57.8 & 71.8 / 76.9 / 89.8 & 56.5 / 60.0 / 75.3 \\
(D) InternVL Fine-tuned \cite{chen2024internvl}	& 72.1 / 75.8 / 75.9 & 56.2 / 57.6 / 57.6 & 72.4 / 77.1 / 89.9 & 56.5 / 61.5 / 76.2 \\
\hline
(E) OED \cite{oed} & 71.5 / 74.5 / 74.5 & 42.1 / 46.1 / 46.1 & 82.3 / 94.6 / 99.0  & 56.4 / 84.0 / 94.6 \\
(F) OED + DSQI (InternVL FT) & 74.8 / 78.0 / 78.0 & \textbf{47.0} / \underline{50.8} / \underline{50.8} & 84.9 / 96.5 / 99.6 & 60.1 / 85.4 / 95.0 \\
(G) OED + DSQI (Qwen FT) & 74.7 / 77.9 / 78.0 & 46.1 / 50.7 / 50.8 & 84.7 / 96.4 / 99.6 & 59.5 / 85.3 / 94.6 \\
(H) OED + DSQI + MMFB (InternVL FT) & \textbf{76.1} / \underline{79.1} / \underline{79.1} & 46.2 / 49.7 / 49.8 & \textbf{85.8} / \underline{96.8} / \underline{99.7} & \underline{60.2} / \textbf{85.7} / \underline{95.1} \\

(I) OED + DSQI + MMFB (Qwen FT) \textbf{(Ours)} & \underline{76.0} / \textbf{79.3} / \textbf{79.3} & \underline{46.5} / \textbf{50.9} / \textbf{50.9} & \underline{85.7} / \textbf{97.0} / \textbf{99.8} & \textbf{63.3} / \underline{85.5} / \textbf{95.3}  \\

\bottomrule
\end{tabular}
}
\caption{Performance of VLM-based methods for spatio-temporal scene graph generation on the Action Genome dataset. For fair comparison with VLMs, we report only spatial context aggregation for (E)-(I), excluding temporal context aggregation. 
DSQI: Dual-Source Query Initialization, MMFB: Multi-Modal Feature Bank, Qwen FT: Qwen Fine-Tuned Embeddings and InternVL FT: InternVL Fine-Tuned Embeddings.}
\label{tab:vlm_performance}
\end{table}

\section{Conclusion}
\label{sec:conclusion}

We present VOST-SGG, a novel framework for spatio-temporal scene graph generation, where we re-think the learnable query design by incorporating semantics from VLMs. 
By disentangling the "what" and "where" components of queries, we ground the queries with meaningful context, leading to improved gains, especially in rare and long-tail predicate scenarios. 
We also propose multi-modal feature bank to further enhances predicate reasoning by integrating visual, textual, and spatial cues into a unified representation. 
In future, we plan to extend this feature bank with pose and motion cues to further enhance relational understanding, and how to extend our framework toward open-ended spatio-temporal scene graph generation.
\newline
\textbf{Acknowledgment} This research/project is supported by the National Research Foundation, Singapore, under its NRF Fellowship (Award\# NRF-NRFF14-2022-0001) and funding allocation to B.F. by A*STAR under its SERC Central Research Fund (CRF).

{
    \small
    \bibliographystyle{ieeenat_fullname}
    \bibliography{main}
}

\clearpage
\setcounter{page}{1}
\setcounter{figure}{5} 
\setcounter{table}{3} 
\setcounter{section}{0} 
\maketitlesupplementary


In this supplementary material, we provide the following additional details and analyses:
(1) the internal architecture of VOST-SGG’s encoders and decoders (Section~\ref{app:encoder_architecture});
(2) the prompts used to fine-tune and extract common sense knowledge priors from VLMs (Section~\ref{app:prompts});
(3) per-predicate scene graph detection performance of VOST-SGG (Section~\ref{app:graphs});
(4) qualitative results (Section~\ref{app:qualtative_results});
(5) ablation study under the SGDET setting (Section \ref{app:ablation});
(6) qualitative results demonstrating VOST-SGG’s Out-of-Distribution (OOD) detection capability on the AVA~\cite{gu2018ava} action detection benchmark (Section \ref{app:ood});
(7) additional results under the SGCLS setting (Section \ref{app:sgcls}); and
(8) efficiency analysis of our model (Section \ref{app:complexity}).
The codebase for our VOST-SGG model will be made publicly available upon acceptance of the paper.

\section{Internal Architecture of Our Encoders and Decoders}
\label{app:encoder_architecture}

\begin{figure}[H]
    \centering
    \includegraphics[width=0.7\linewidth]{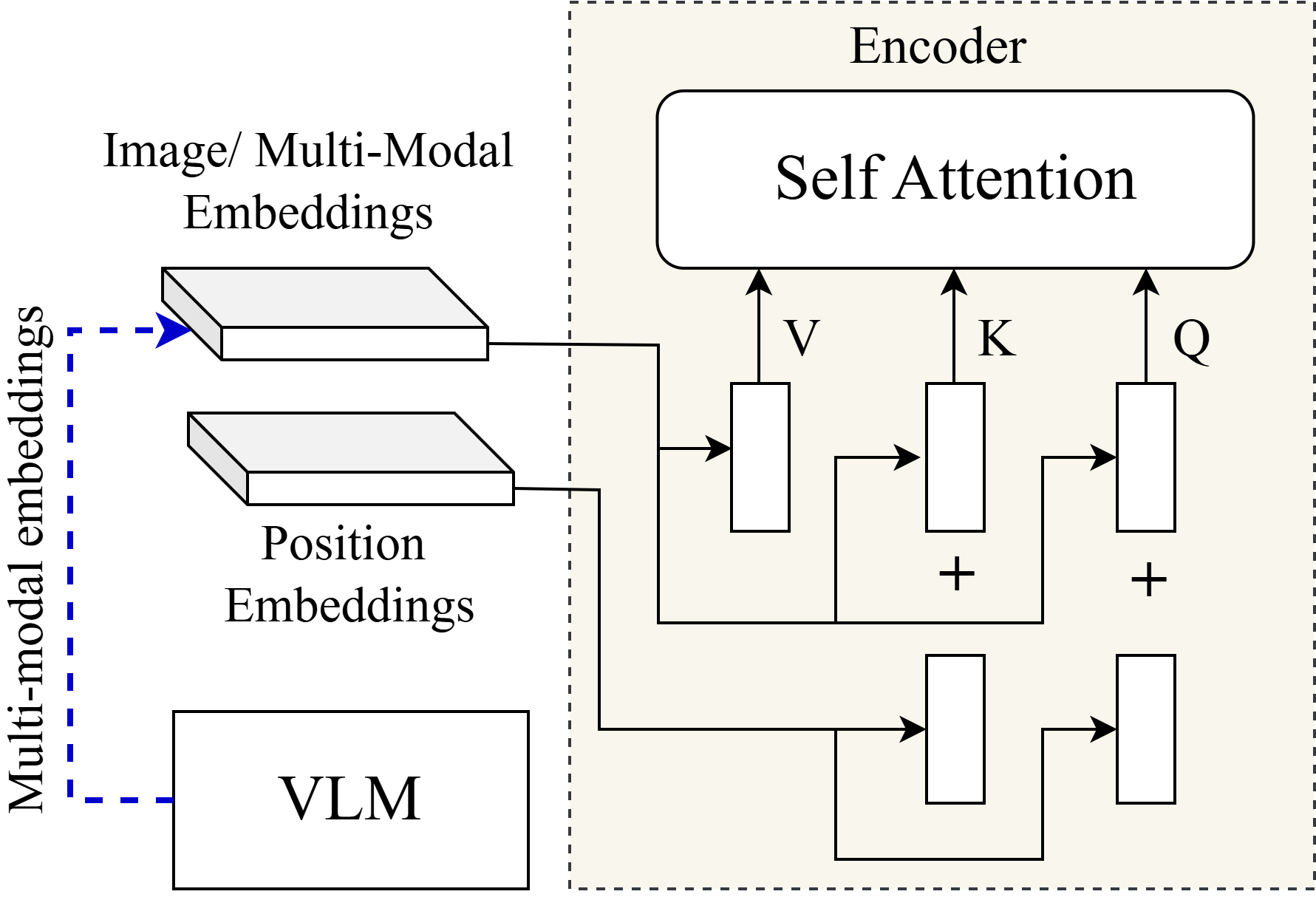}
    \caption{Encoder architecture.}
    \label{fig:encoder_internal}
\end{figure}

\begin{figure}[t]
    \centering
    \includegraphics[width=0.7\linewidth]{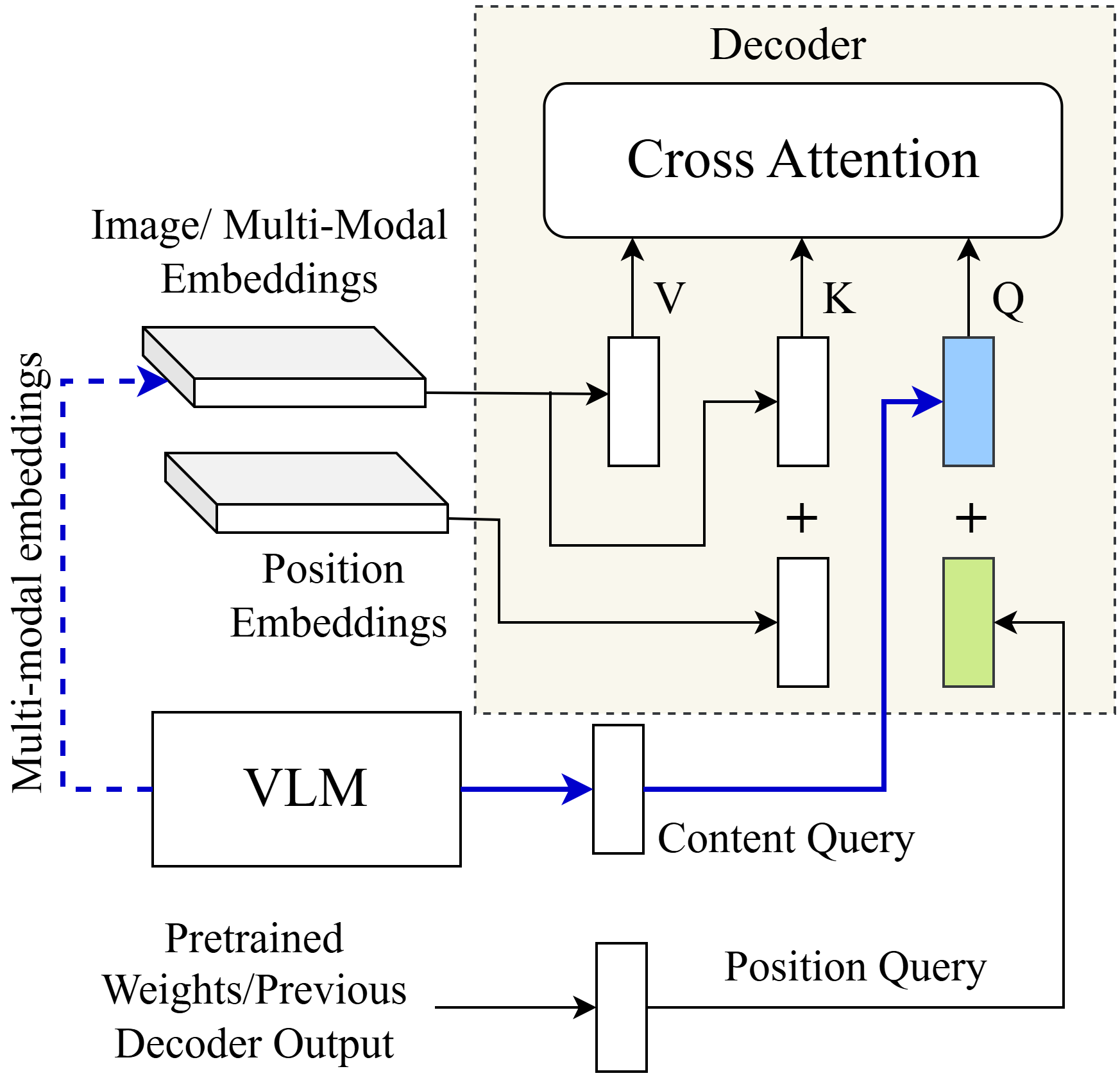}
    \caption{Decoder architecture.}
    \label{fig:decoder_internal}
\end{figure}

The internal architecture of our encoders and decoders are illustrated in Fig. \ref{fig:encoder_internal} and Fig. \ref{fig:decoder_internal} respectively. 
Both the image and multi-modal encoders utilize sinusoidal positional embeddings.
In the multi-modal encoder, elements of \( \mathcal{F}^{\text{MMBank}}\) in Eq. 9
are concatenated along the sequence dimension before being processed by the encoder.

\section{VLM Prompts}
\label{app:prompts}

\begin{figure}
    \centering
    \begin{subfigure}[]{1.0\linewidth}
        \centering
        \includegraphics[width=1.0\linewidth]{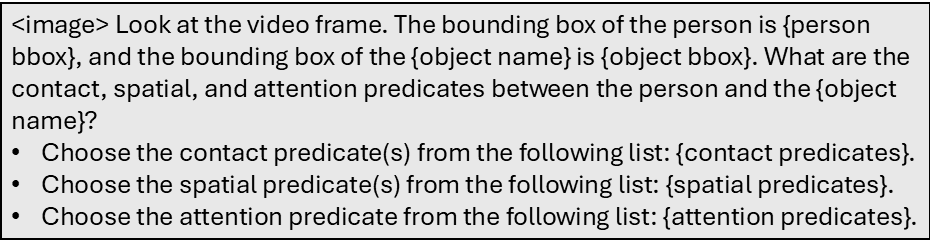}
        \subcaption{PredCLS prompt.}
        \label{}
    \end{subfigure}
    \hfill
    \begin{subfigure}[]{1.0\linewidth}
        \centering
        \includegraphics[width=1.0\linewidth]{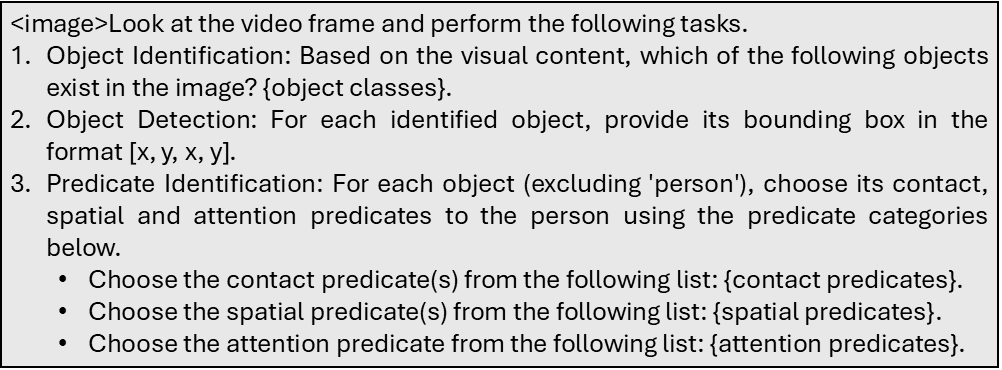}
        \subcaption{SGDET prompt.}
        \label{}
    \end{subfigure}
    \caption{VLM prompts.}
    \label{fig:qwen_prompts}
\end{figure}

We use the prompt templates shown in Fig.~\ref{fig:qwen_prompts} for both VLM fine-tuning and response extraction (in both the zero-shot and fine-tuned settings). Following the fine-tuning protocol described in~\cite{zhao2025swift}, we fine-tune the Qwen2.5-VL-7B-Instruct model~\cite{bai2025qwen2} on the Action Genome~\cite{ag} training set for one epoch. During fine-tuning, we employ the LoRA technique, which updates only a small subset of 20M parameters, compared to the 7.6B parameters of the full model, ensuring parameter-efficient adaptation. Fine-tuning was performed separately for the SGDET and predCLS settings.


\section{Per-Predicate Scene Graph Detection Performance}
\label{app:graphs}

\begin{figure}
    \includegraphics[width=\linewidth]{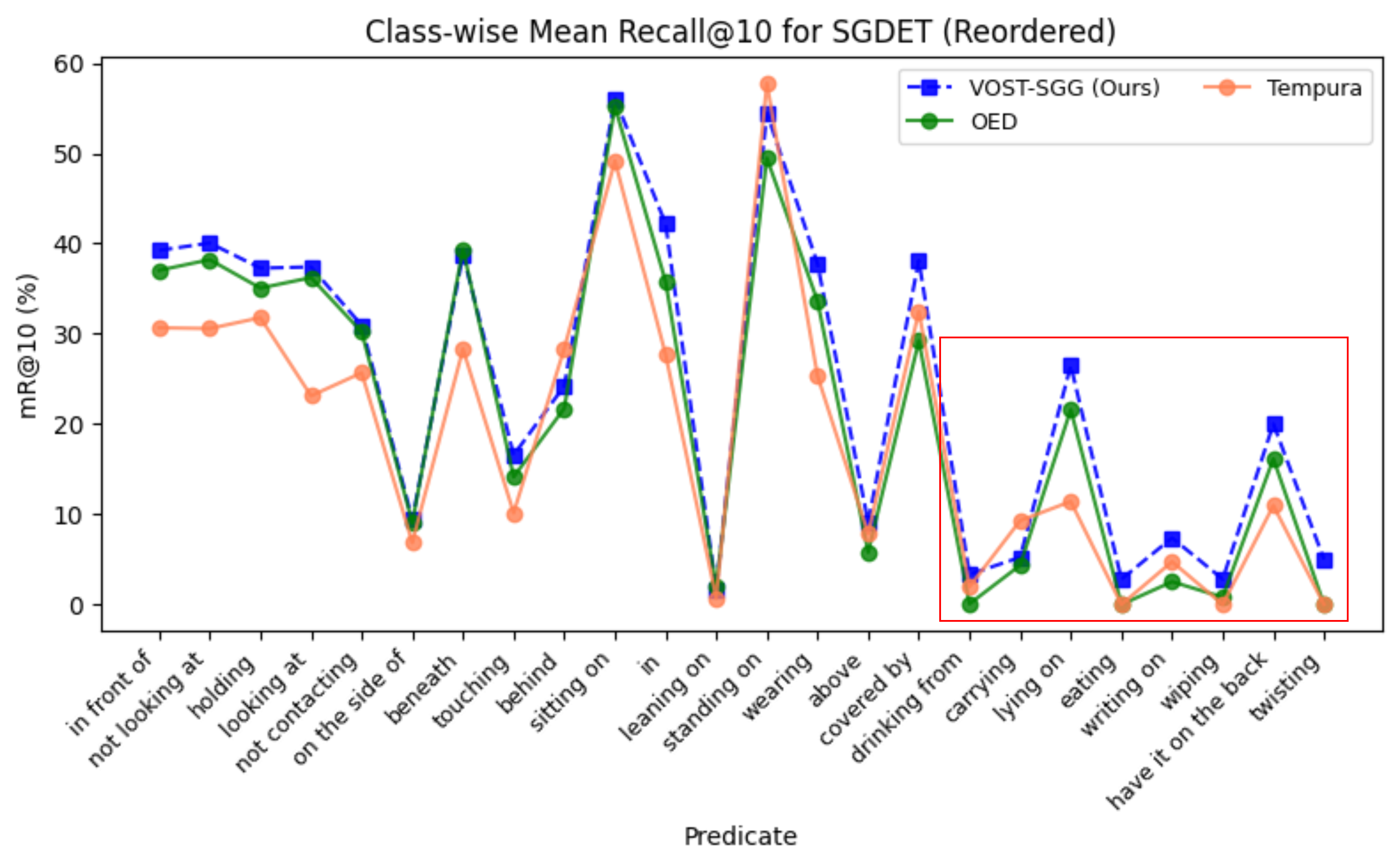}
    \caption{Per-predicate mR@10 (SGDET, with constraint, test set). We compare VOST-SGG to the single-stage OED~\cite{oed} and multi-stage TEMPURA~\cite{tempura}. Predicates are sorted by frequency in descening order from left to right.}
    \label{fig:mR_line_sgdet}
    \label{}
\end{figure}

In Fig. \ref{fig:mR_line_sgdet}, we present a per-class performance comparison of our VOST-SGG against prior single-stage (OED~\cite{oed}) and multi-stage (TEMPURA~\cite{tempura}) models under the mR@10 metric for the SGDET (with constraint) setting.
VOST-SGG consistently achieves higher performance across most predicate classes, highlighting its ability to capture diverse predicates effectively. While TEMPURA \cite{tempura}, despite being designed to address long-tail distributions, underperforms compared to both OED \cite{oed} and VOST-SGG, particularly on rare predicates.
Notably, even though VOST-SGG modestly outperforms the baselines on head classes, it shows significant improvements on both body (mid frequency) and tail (low frequency) classes. Table \ref{tab:predicate_counts} shows, the frequency of predicates in Action Genome dataset. This highlights the strength of VOST-SGG’s query re-design and multi-modal reasoning, especially in modelling rare relationships, resulting in more balanced and comprehensive scene graph generation.

\begin{table}[t]
\centering
\resizebox{0.7\linewidth}{!}{
\begin{tabular}{|l|c|c|c|}
\toprule
\textbf{Predicate} & \textbf{All} & \textbf{Train} & \textbf{Test} \\
\midrule
in front of           & 384,913 & 277,032 & 107,881 \\
not looking at        & 300,084 & 212,640 & 87,444  \\
holding               & 234,929 & 169,741 & 65,188  \\
looking at            & 217,435 & 155,780 & 61,655  \\
not contacting        & 164,043 & 117,088 & 46,955  \\
on the side of        & 111,363 & 79,004  & 32,359  \\
beneath               & 93,511  & 65,347  & 28,164  \\
touching              & 78,933  & 57,033  & 21,900  \\
behind                & 74,922  & 52,769  & 22,153  \\
unsure                & 65,806  & 48,882  & 16,924  \\
sitting on            & 62,100  & 43,633  & 18,467  \\
in                    & 22,687  & 16,154  & 6,533   \\
leaning on            & 17,766  & 12,944  & 4,822   \\
other relationship    & 14,406  & 10,084  & 4,322   \\
standing on           & 13,107  & 9,124   & 3,983   \\
wearing               & 10,524  & 7,611   & 2,913   \\
above                 & 9,137   & 6,116   & 3,021   \\
lying on              & 8,921   & 6,330   & 2,591   \\
covered by            & 8,346   & 5,705   & 2,641   \\
carrying              & 6,479   & 4,430   & 2,049   \\
drinking from         & 6,102   & 4,577   & 1,525   \\
eating                & 4,589   & 3,355   & 1,234   \\
writing on            & 1,443   & 1,159   & 284     \\
wiping                & 1,131   & 842     & 289     \\
have it on the back   & 536     & 358     & 178     \\
twisting              & 121     & 93      & 28      \\
\bottomrule
\end{tabular}
}
\caption{Predicate frequencies across splits (All/Train/Test), ordered by the All column in descending order.}
\label{tab:predicate_counts}
\end{table}




\section{Qualitative Results}
\label{app:qualtative_results}

In this section, we present qualitative results of VOST-SGG’s predictions. 
In Section~\ref{app:AG_examples}, we compare our model with OED~\cite{oed} on the Action Genome dataset. In Section~\ref{app:Qwen_examples}, we further compare the original VLM responses produced by Qwen \cite{bai2025qwen2} with the final VOST-SGG outputs, illustrating how VOST-SGG effectively leverages VLM knowledge, through its two key architectural components, to generate more reliable spatio-temporal scene graphs.

\subsection{Qualitative comparison of ST-SGG predictions between our VOST-SGG and OED}
\label{app:AG_examples}

\begin{figure}[!t]
    \centering
    \begin{subfigure}[]{1.0\linewidth}
        \centering
        \includegraphics[width=\linewidth]{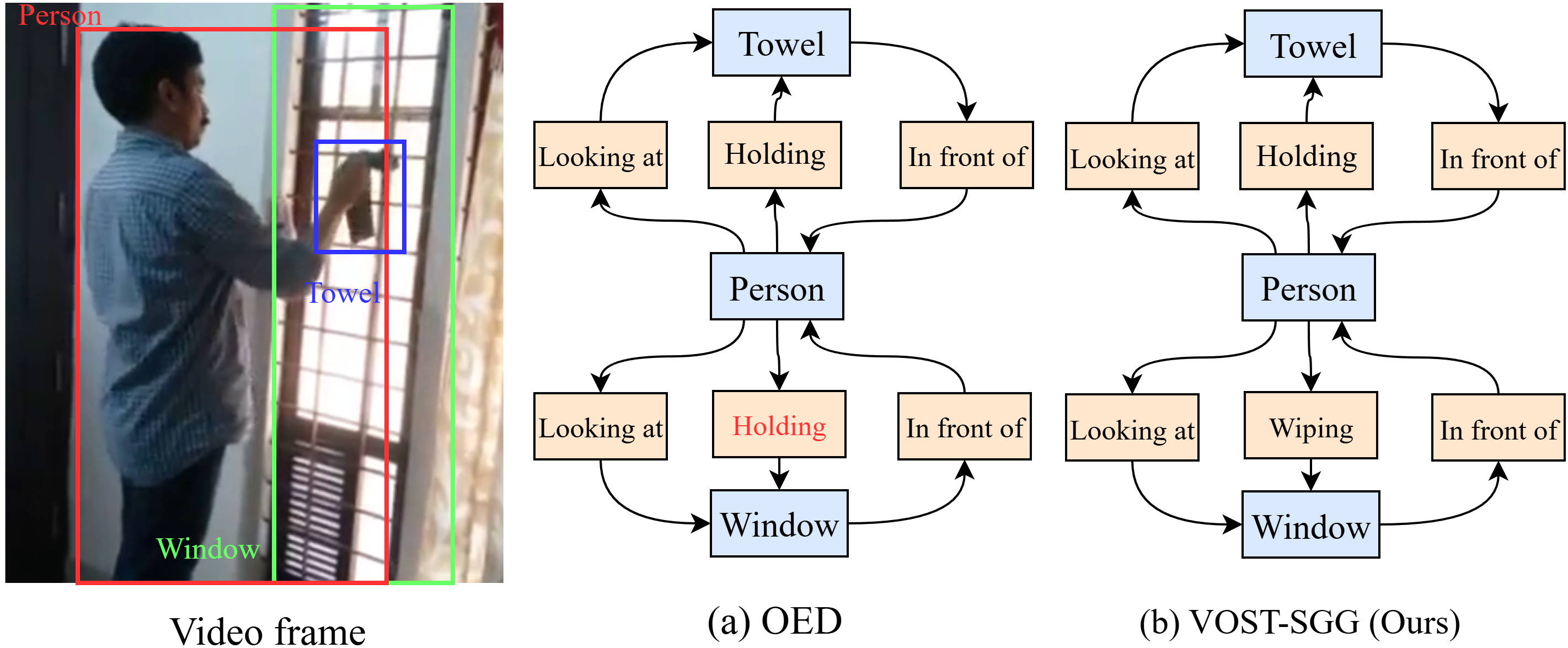}
        \subcaption{}
        \label{fig:qualitative_results_main}
    \end{subfigure}
    \hfill
    \begin{subfigure}[]{1.0\linewidth}
        \centering
        \includegraphics[width=\linewidth]{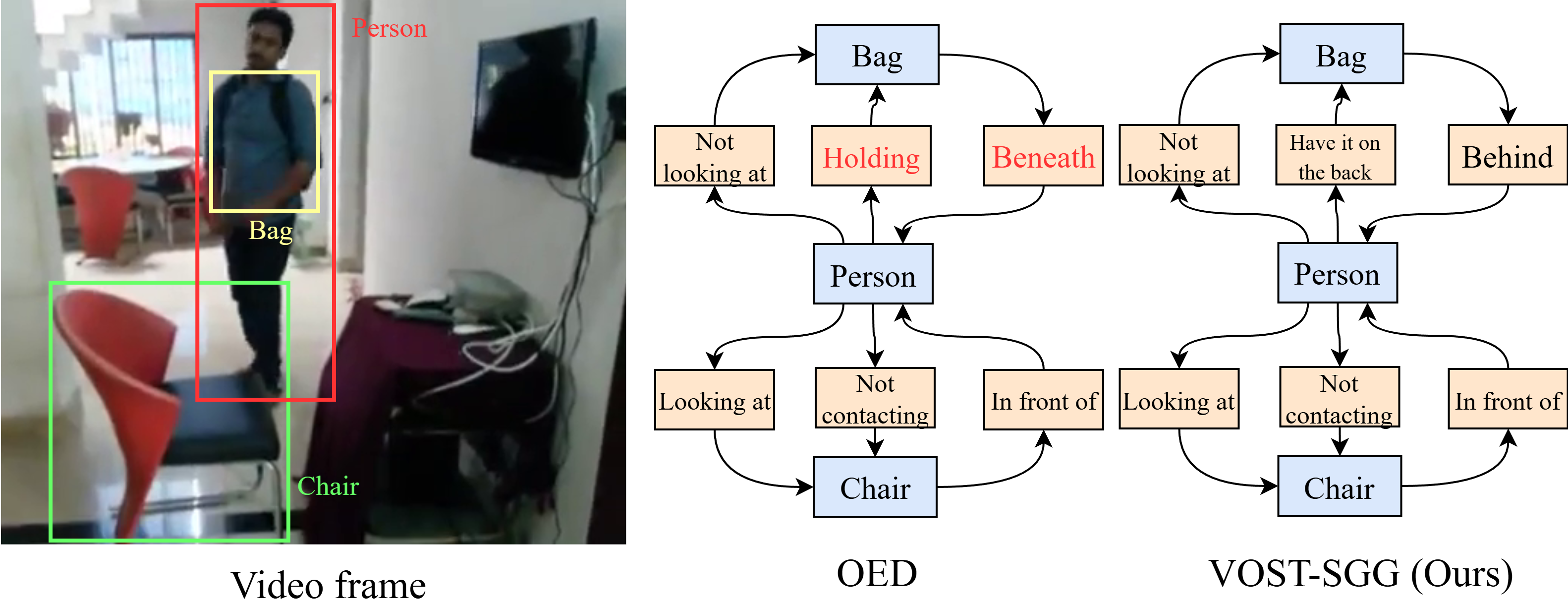}
        \subcaption{}
        \label{fig:qualitative_results_main2}
    \end{subfigure}
    \caption{Qualitative comparison between VOST-SGG (ours) and OED \cite{oed} on spatio-temporal scene graph generation \textbf{predCLS} (with constraint) task.}
    \label{fig:qualitative_results_pred_cls}
\end{figure}

\begin{figure}[!t]
    \centering
    \begin{subfigure}[]{1.0\linewidth}
        \centering
        \includegraphics[width=\linewidth]{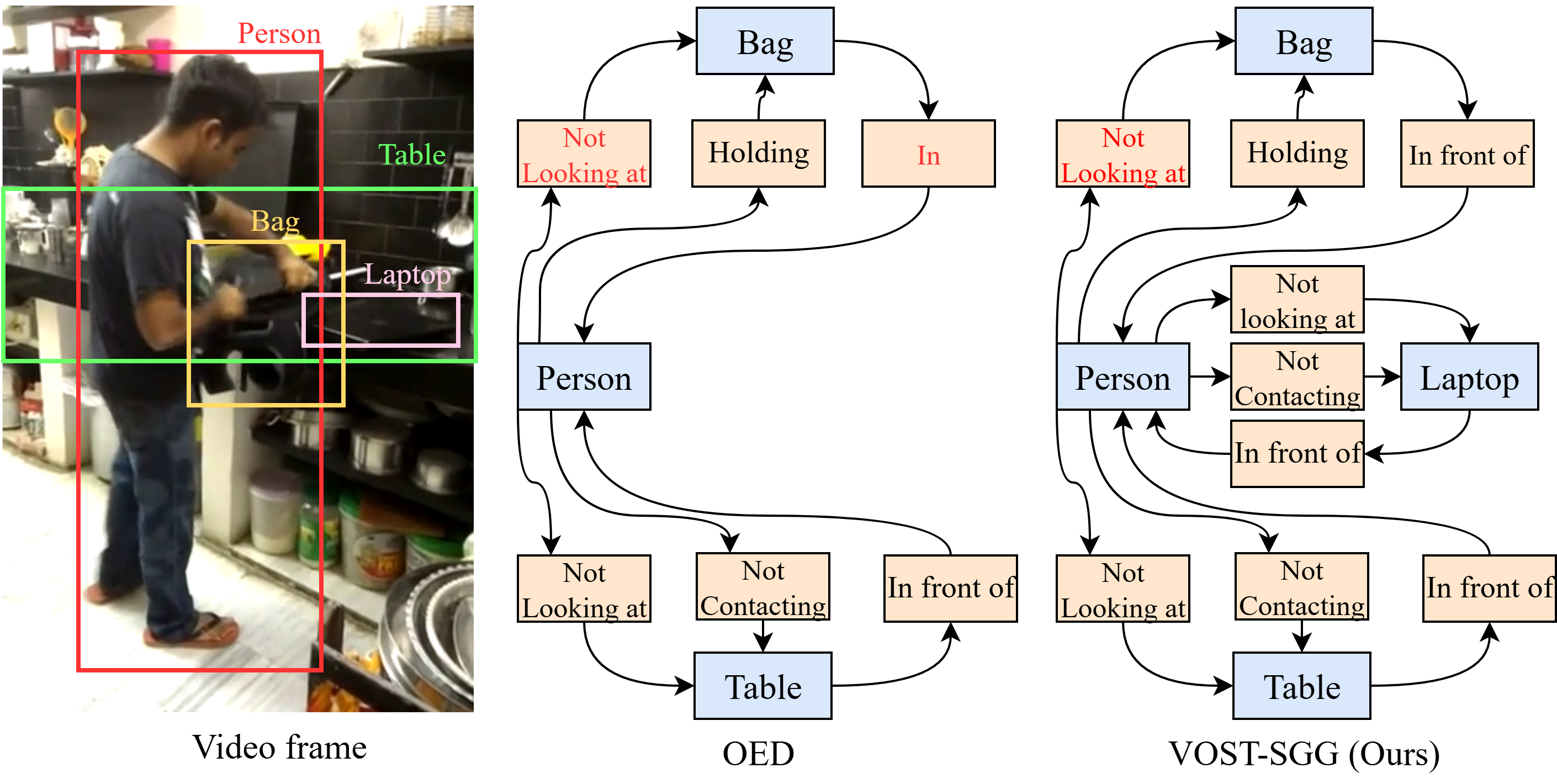}
        \subcaption{}
        \label{fig:missed_obj}
    \end{subfigure}
    \hfill
    \begin{subfigure}[]{1.0\linewidth}
        \centering
        \includegraphics[width=\linewidth]{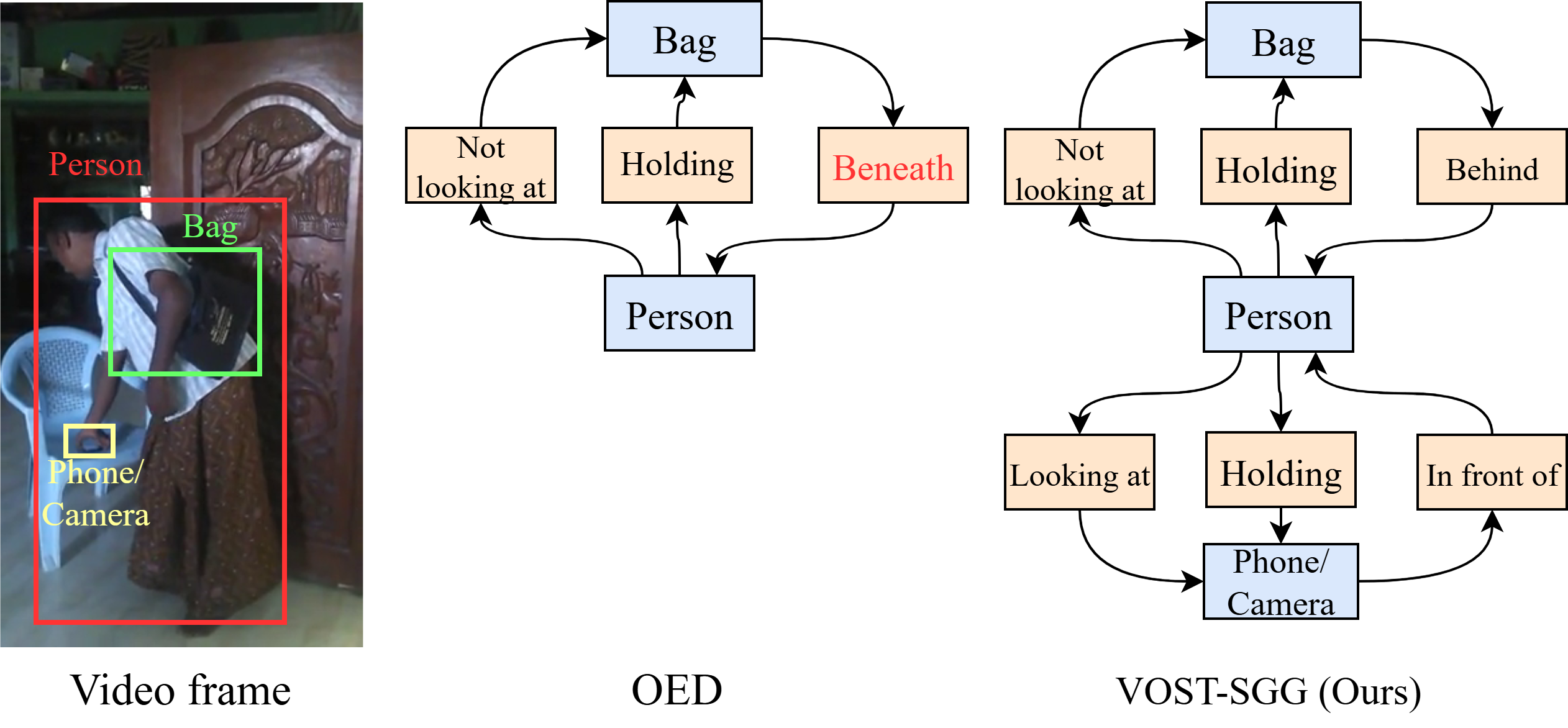}
        \subcaption{}
        \label{fig:small_obj}
    \end{subfigure}
    \caption{Qualitative comparison between VOST-SGG (ours) and OED \cite{oed} on spatio-temporal scene graph generation \textbf{SGDET} (with constraint) task.}
    \label{fig:qualitative_results}
\end{figure}

In \ref{fig:qualitative_results_pred_cls} and Fig. \ref{fig:qualitative_results}, we provide qualitative comparison of ST-SGG predictions between our VOST-SGG and OED \cite{oed} under predCLS (with constraint) and SGDET (with constraint) settings respectively.

\subsection{Effective Use of VLM priors through Dual-Source Query Initialization and Multi-Modal Feature Bank}
\label{app:Qwen_examples}

In this section, we demonstrate how our model leverages VLM knowledge through our key architectural innovations dual-source query initialization and a multi-modal feature bank to produce more reliable spatio-temporal scene graphs.
We compare VOST-SGG’s predictions with the initial responses generated by Qwen (Qwen-FT) under the SGDET (with constraints) setting. For a fair comparison with Qwen-FT, we disable the temporal aggregation module in VOST-SGG during this analysis.

\begin{figure}[!t]
    \centering
    \begin{subfigure}[]{1.0\linewidth}
        \centering
        \includegraphics[width=\linewidth]{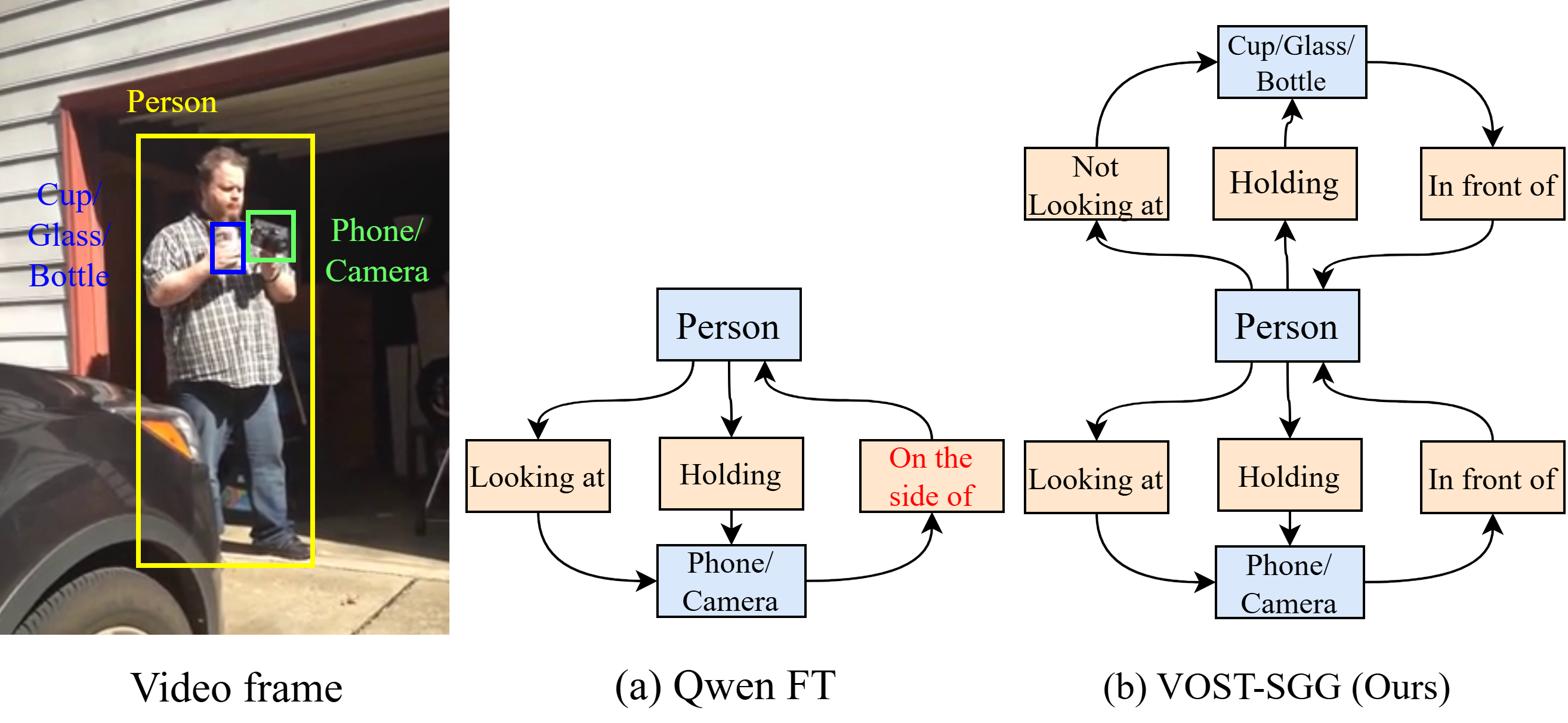}
        \subcaption{}
        \label{fig:qwen_correct_obj}
    \end{subfigure}
    \hfill
    
    \begin{subfigure}[]{1.0\linewidth}
        \centering
        \includegraphics[width=\linewidth]{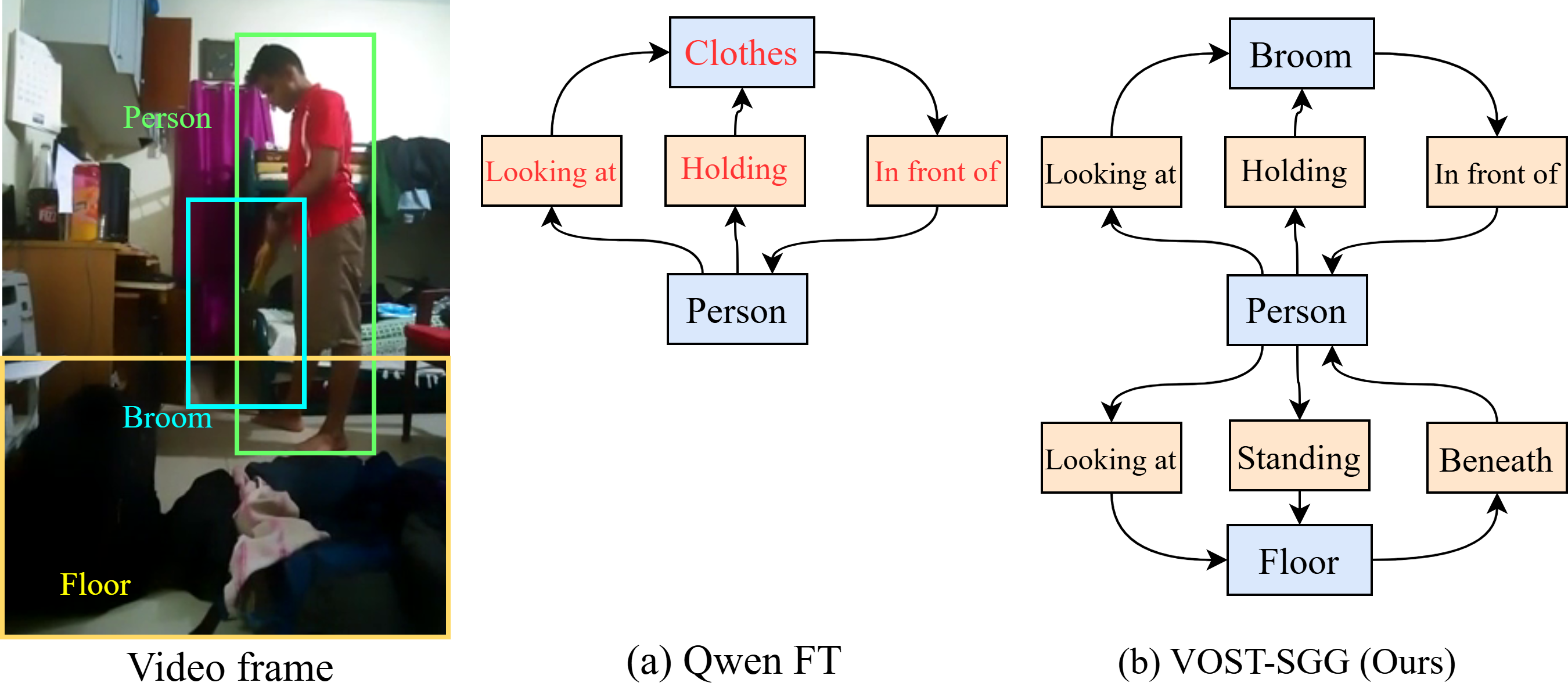}
        \subcaption{}
        \label{fig:qwen_incorrect_obj}
    \end{subfigure}
    \hfill
    
    \begin{subfigure}[]{1.0\linewidth}
        \centering
        \includegraphics[width=\linewidth]{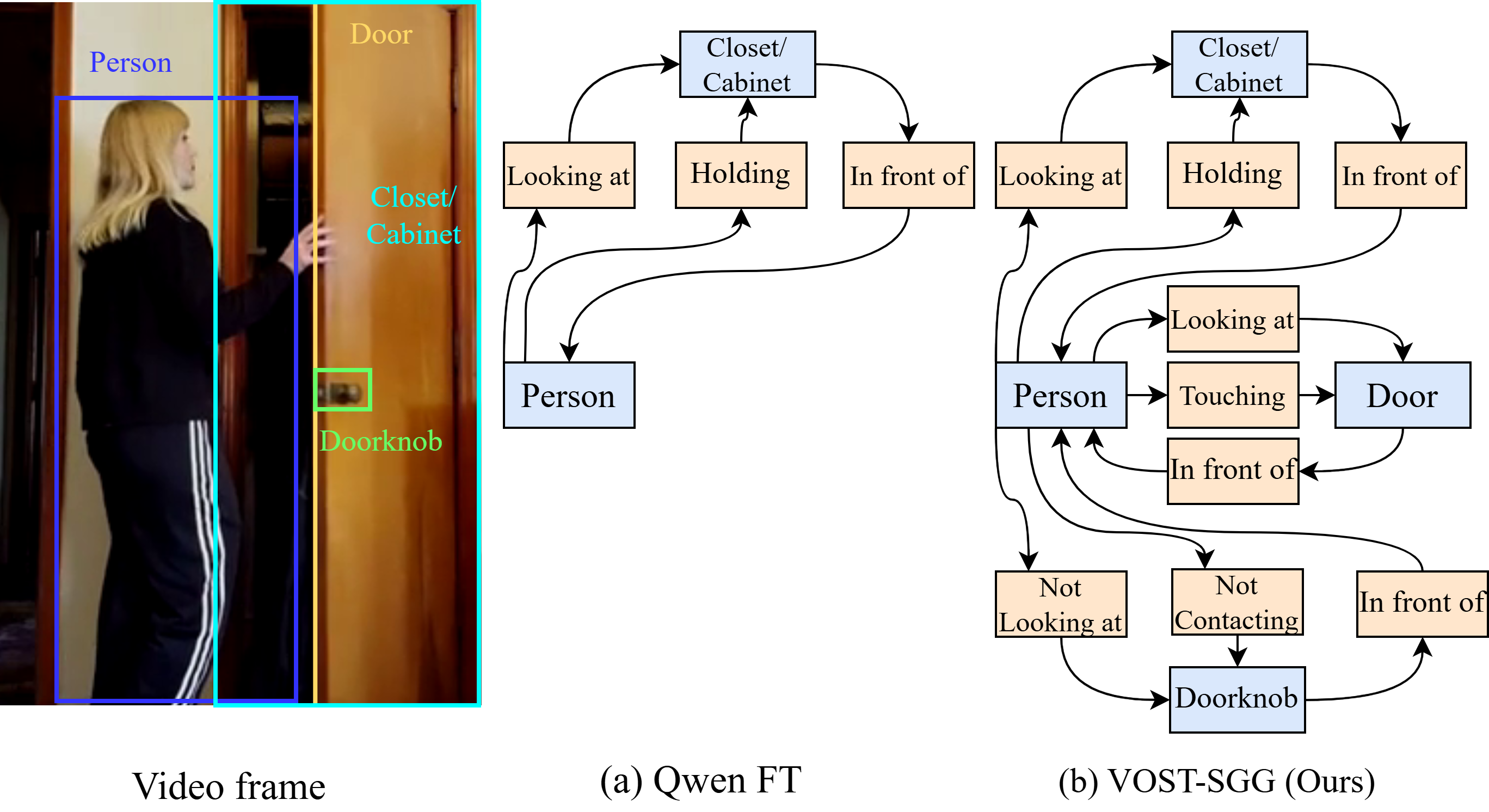}
        \subcaption{}
        \label{fig:qwen_context}
    \end{subfigure}
 
    \caption{Qualitative illustration of VOST-SGG’s ability to effectively utilize VLM priors for spatio-temporal scene graph generation. 
    Qwen FT refers to the Qwen fine-tuned model on Action Genome \cite{ag} dataset which we used to extract the initial VLM responses.
    Results are presented under the \textbf{SGDET} (with constraints) evaluation setting.}
    \label{fig:qwen_qualitative_results}
\end{figure}

In Fig. \ref{fig:qwen_correct_obj}, Qwen correctly identifies one interacting object, \emph{Phone/Camera}, along with the associated predicates. VOST-SGG successfully incorporates this correct prior while additionally detecting the second interacting object, \emph{Cup/Glass/Bottle} and inferring its predicates, which were entirely missed by Qwen while also correcting one of Qwen’s initial predicate mispredictions.
Further, in Fig. \ref{fig:qwen_incorrect_obj}, the initial VLM prediction \emph{Clothes} is incorrect. Although \emph{Clothes} are present on the 
\emph{Floor}, the person is not interacting with them; instead, they are cleaning the \emph{Floor} with a \emph{Broom}. Despite this incorrect VLM cue, VOST-SGG suppresses the misprediction and correctly identifies the \emph{Floor} and \emph{Broom} as the interacting objects, along with the appropriate predicates.
Moreover, in Fig. \ref{fig:qwen_context}, the VLM prediction partially captures the scene by identifying the \emph{Closet/Cabinet} but fails to detect its \emph{Door} and \emph{Doorknob}. In this case, VOST-SGG is probably leveraging the contextual prior provided by the VLM, the presence of a \emph{Closet/Cabinet}, to recover the missing objects, including the fine-grained \emph{Doorknob}, and correctly infer their predicates.

Overall, these examples show that our model does not rely on VLM predictions as rigid supervision. Instead, it uses them as soft priors: adapting them when correct, ignoring them when incorrect, and leveraging their contextual cues when partially correct to infer missing or more fine-grained interactions.

\section{Ablation Study for SGDET Setting}
\label{app:ablation}

The ablation study for SGDET setting, evaluated under with and no constraint scenarios using Recall and mean Recall metrics, is presented in Tab.~\ref{tab:app:ablation}.

\begin{table}[ht]
\centering
\resizebox{\columnwidth}{!}{
\begin{tabular}{|l|c|c|c|c|}
\toprule

\multirow{2}{*}{Method} & \multicolumn{2}{c|}{With Constraint} & \multicolumn{2}{c|}{No Constraint}   \\
\cline{2-5}
&   R@10 / 20 / 50 & mR@10 / 20 / 50 & R@10 / 20 / 50 & mR@10 / 20 / 50 \\

\toprule

IAQI  & 33.5 / 40.9 / \underline{48.9} & 20.8 / 26.5 / 32.5 & 35.3 / 44.0 / \underline{51.8} & 
\textbf{32.6} / 39.1 / \underline{49.7} \\

DSQI	& \underline{33.7} / \underline{41.2} / \textbf{49.0}	& \underline{22.3} / \underline{28.8} / \underline{34.6}	& \underline{35.7} / \underline{44.5} / \textbf{52.6}	& 28.4 / \underline{40.5} / \textbf{51.5} \\

MMFB   & \textbf{34.2} / \textbf{41.5} / \textbf{49.0} & \textbf{23.0} / \textbf{29.7} / \textbf{35.8} & \textbf{36.5} / \textbf{44.8} / \textbf{52.6} & \underline{28.7} / \textbf{41.6} / \textbf{51.5} \\

\bottomrule
\end{tabular}
}
\caption{
Ablation study on Action Genome dataset \cite{ag} for query re-design and multi-modal feature bank for \textbf{SGDET} setting. 
IAQI: Instance-Agnostic Query Initialization, 
DSQI: Dual-Source Query Initialization, 
MMFB: Multi-Modal Feature Bank. 
The bold and underline font shows the best and the second best result, respectively.}
\label{tab:app:ablation}
\end{table}

\section{Inference of VOST-SGG on Out-of-Distribution Datasets}
\label{app:ood}

We evaluate the Out-of-Distribution (OOD) detection capability of VOST-SGG using some test samples of AVA \cite{gu2018ava} dataset’s action detection benchmark. 
Among the 80 action classes in AVA, nine Action Genome \cite{ag} contact-predicate classes \emph{Sitting, Lying, Sleeping, Eating, Carrying, Holding, Writing, Standing, Touching} directly overlap with AVA’s action classes. 
Although a single person instance may be associated with multiple actions in the same frame, AVA treats each action as an independent test sample. In contrast, since VOST-SGG can predict multiple predicates associated with one person per frame, we treat all annotated actions for a given person instance in that frame as the ground-truth set the model must recover. Test samples were selected based on whether their AVA action annotations overlap with the Action Genome predicates.

\begin{figure}[!t]
    \centering

    \begin{subfigure}{0.48\linewidth}
        \centering
        \includegraphics[width=\linewidth]{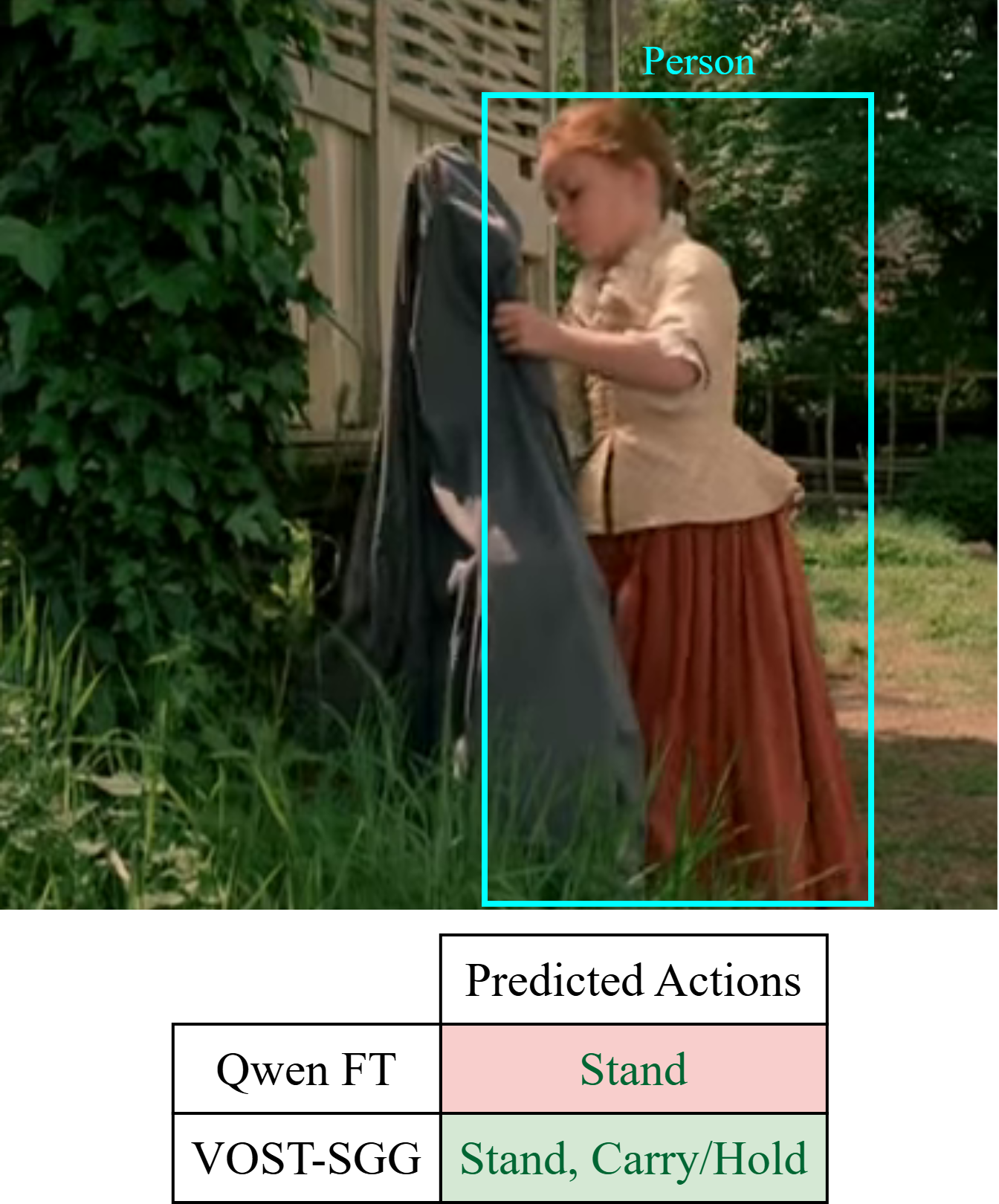}
        \subcaption{}
        \label{fig:ava_carry_stand}
    \end{subfigure}
    \hfill
    \begin{subfigure}{0.48\linewidth}
        \centering
        \includegraphics[width=\linewidth]{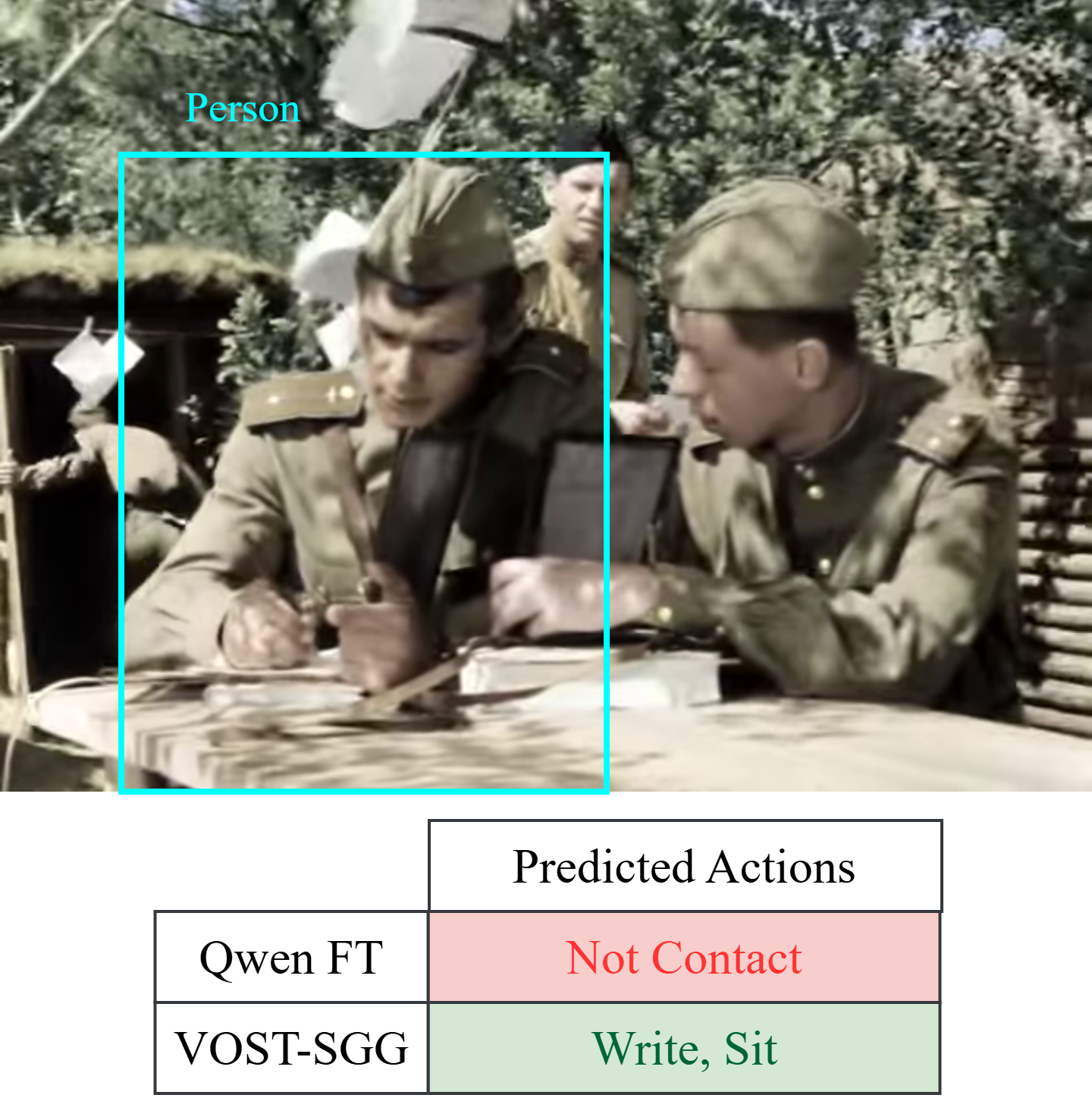}
        \subcaption{}
        \label{fig:ava_sit_write}
    \end{subfigure}

    \vspace{6pt}

    \begin{subfigure}{0.48\linewidth}
        \centering
        \includegraphics[width=\linewidth]{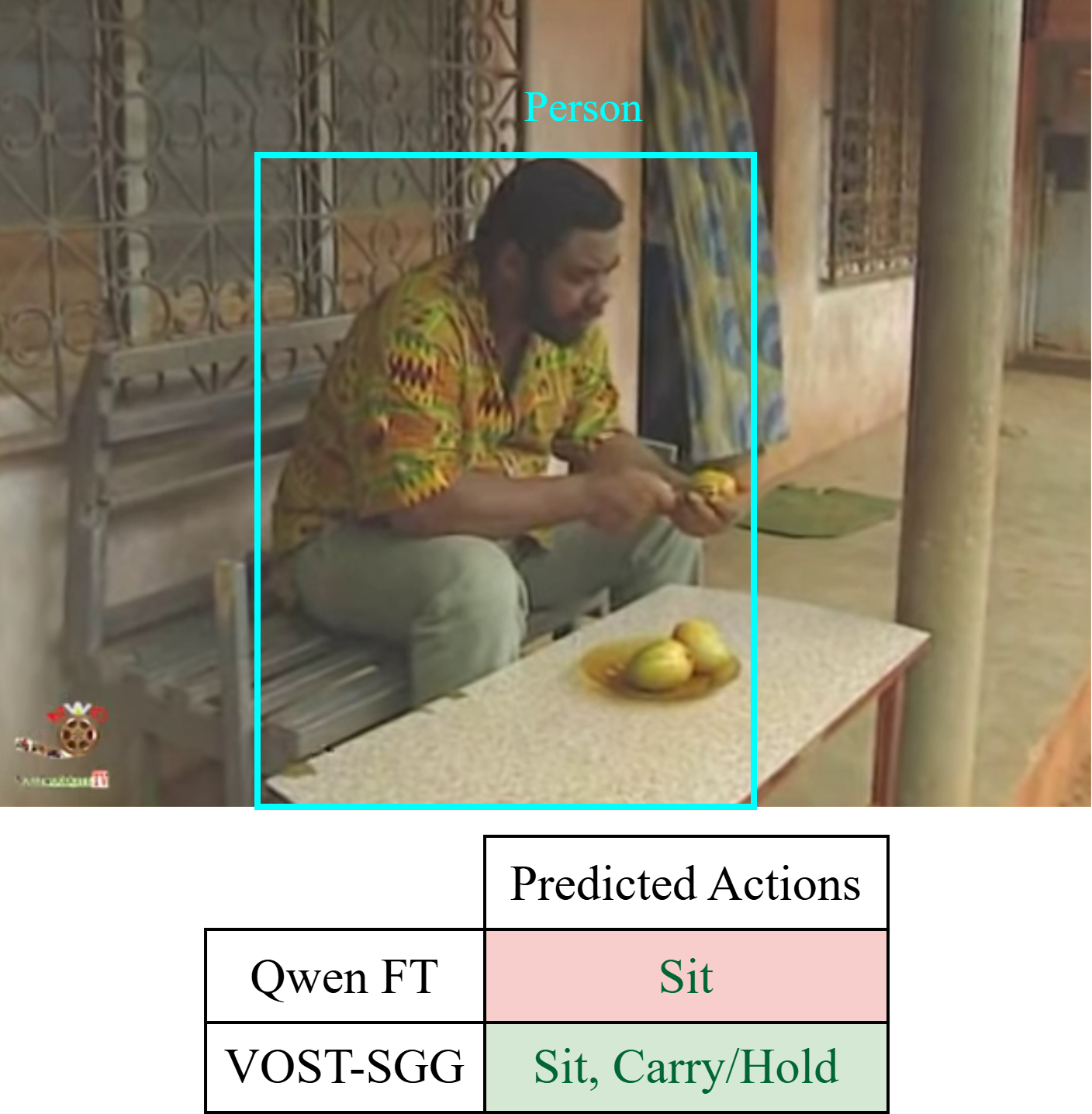}
        \subcaption{}
        \label{fig:ava_sit_hold}
    \end{subfigure}
    \hfill
    \begin{subfigure}{0.48\linewidth}
        \centering
        \includegraphics[width=\linewidth]{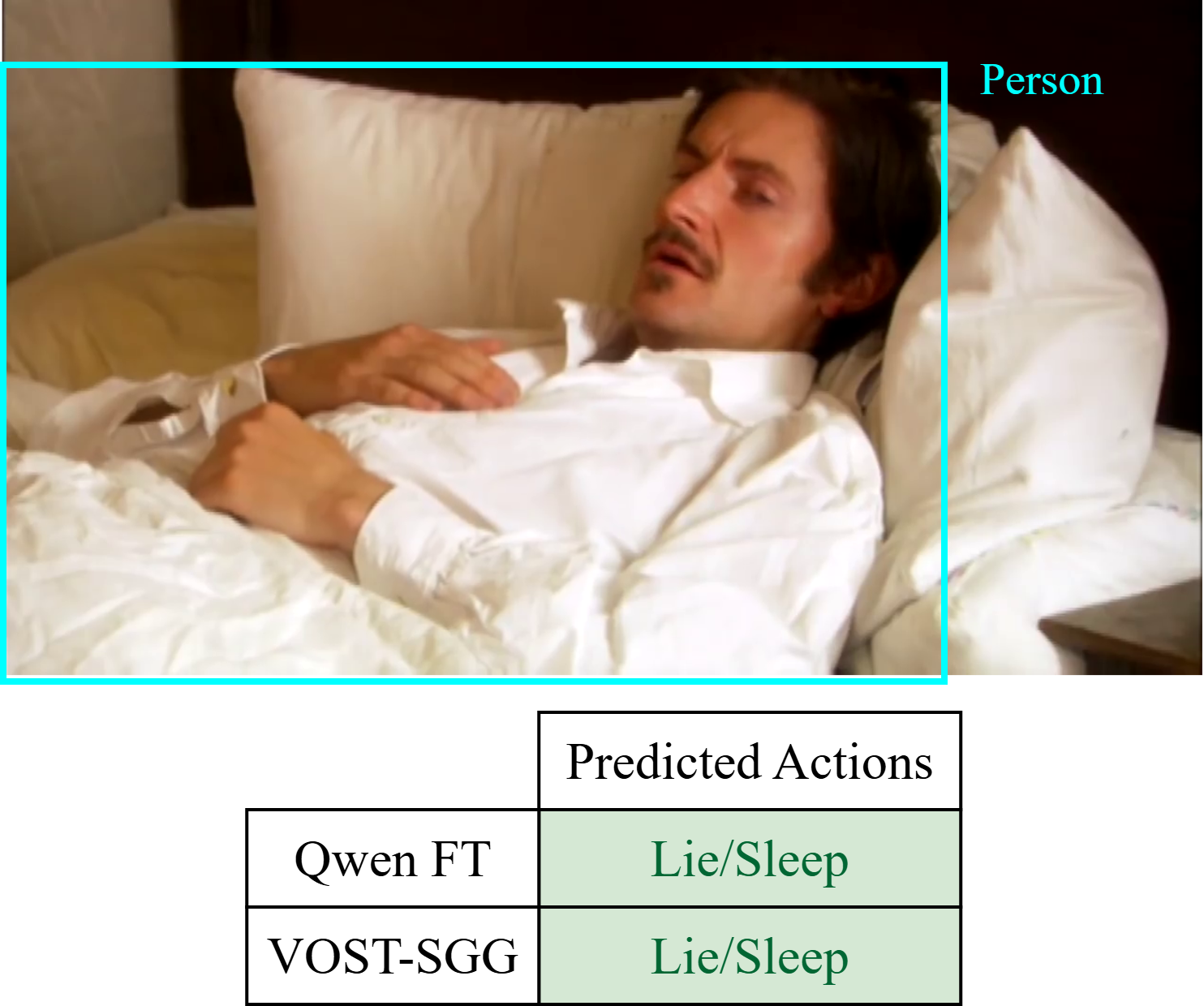}
        \subcaption{}
        \label{fig:ava_sleep}
    \end{subfigure}

    \caption{Qualitative demonstration of VOST-SGG’s OOD generalization on the AVA dataset. Correct action predictions are shown in green font, incorrect predictions in red. Prediction text boxes are shaded green or red depending on whether all annotated actions for that sample are correctly identified by the model.}
    \label{fig:ava_ood}
\end{figure}

In Fig. \ref{fig:ava_ood}, we compare VOST-SGG with the fine-tuned Qwen model (Qwen-FT) under the SGDET setting. For fair comparison with Qwen FT model, we disable the temporal aggregation module from VOST-SGG.
%
The correct action predictions are shown in green font, incorrect predictions in red.
The prediction text boxes are shaded green or red depending on whether all annotated actions for that frame were correctly identified by the model. 

As illustrated, VOST-SGG consistently identifies all relevant action classes, whereas Qwen-FT often partially identifies the correct actions and additionally produces incorrect predictions, such as in Fig.~\ref{fig:ava_sit_write}.
These results indicate that VOST-SGG has the capability to generalize effectively to OOD video data delivering reliable action-level predictions on datasets outside its training domain.

\section{Additional Results Under SGCLS Setting}
\label{app:sgcls}

We also evaluate our model under the Scene Graph Classification (SGCLS) setting, as shown in Table~\ref{tab:sgcls}. In the SGCLS setting, the ground-truth bounding boxes are provided, and the model should correctly classify the ground truth bounding box classes and predict predicate labels. Consistent with the results in the PREDCLS and SGDET settings, VOST-SGG outperforms all the prior models by a significant margin.

\begin{table}[t]
\centering
\resizebox{0.7\columnwidth}{!}{
\begin{tabular}{|l|c|c|}
\toprule

\multirow{2}{*}{Method} & \multicolumn{1}{c|}{With Constraint} & \multicolumn{1}{c|}{No Constraint}   \\
\cline{2-3}
&   R@10 / 20 / 50  & R@10 / 20 / 50  \\

\toprule

VRD \cite{zheng2022vrdformer}       & 32.4 / 33.3 / 33.3 &  32.4 / 33.3 / 33.3 \\
M-FREQ \cite{zellers2018neural}     & 40.8 / 41.9 / 41.9 &  40.8 / 41.9 / 41.9 \\
VCTree \cite{tang2019learning}      & 44.1 / 45.3 / 45.3 &  44.1 / 45.3 / 45.3 \\ 
RelDN \cite{zhang2019graphical}     & 11.0 / 11.0 / 11.0 & 25.0 / 41.9 / 47.9 \\
GPS-Net \cite{lin2020gps}           & 45.3 / 46.5 / 46.5 & - \\

\hline

TRACE \cite{trace}                  & 14.8 / 14.8 / 14.8  & 37.1 / 46.7 / 50.5 \\
STTran \cite{sttran}                & 46.4 / 47.5 / 47.5  & 54.0 / 63.7 / 66.4 \\
APT \cite{apt}                      & 47.2 / 48.9 / 48.9 &   55.1 / 65.1 / 68.7  \\
STTran-TPI \cite{wang2022dynamic}   & 47.2 / 48.3 / 48.3 &   - \\
TR2 \cite{tr2}                      & 47.7 / 48.7 / 48.7 &  57.2 / 64.4 / 66.2 \\
TEMPURA \cite{tempura}              & 47.2 / 48.3 / 48.3 &  56.3 / 64.7 / 67.9 \\
DSG-DETR \cite{feng2023exploiting}  & 50.8 / 52.0 / 52.0 &   59.2 / 69.1 / 72.4 \\
DIFFVSGG \cite{chen2025diffvsgg}    & \underline{52.5} / \underline{53.7} / \underline{53.7} & \underline{60.5} / \underline{70.5} / \underline{74.4} \\

\hline
OED \cite{oed}                       & 40.1 / 41.4 / 41.5 & 46.7 / 52.7 / 55.2  \\
VOST-SGG (ours)                & \textbf{59.7} / \textbf{61.3} / \textbf{61.4} & \textbf{68.4} / \textbf{75.8} / \textbf{78.2}  \\

\bottomrule
\end{tabular}
}
\caption{
Comparison with state-of-the-art spatio-temporal scene graph generation methods under \textbf{SGCLS} setting on Action Genome dataset \cite{ag}. 
The bold and underline font shows the best and the second best result, respectively.}
\label{tab:sgcls}
\end{table}

\section{Efficiency Analysis}
\label{app:complexity}

DETR-style models are known to require substantially higher training time \cite{liu2022dab}.
Since our approach builds on top of the OED framework \cite{oed}, the overall training process additionally includes VLM fine-tuning and a response extraction phase. The latter incurs no repeated cost, as the extracted VLM outputs are cached and reused throughout training. 
The fine-tuning stage required approximately 32 hours on a single 49 GB NVIDIA RTX A6000 GPU.

%
Under the PREDCLS and SGCLS settings, VOST-SGG achieves a latency of 11.3 ms/frame while OED requires 73.7 ms/frame. Under the more demanding SGDET setting, OED attains 93.7 ms/frame, whereas VOST-SGG reduces this to 28.6 ms/frame.
In terms of model capacity, our PREDCLS and SGCLS variants contain 54.97M parameters, compared to 54.73M in OED. For SGDET, VOST-SGG has 69.3M parameters, whereas OED uses 64.02M. Importantly, despite these differences in total parameter count, the number of trainable parameters remains the same for both VOST-SGG and OED models as 12.89M, due to the use of lightweight reasoning and decoding modules.

\end{document}